\documentclass[10pt,letterpaper]{article}
\usepackage[top=0.85in,left=2.75in,footskip=0.75in]{geometry}

\usepackage{amsmath,amssymb}

\usepackage{changepage}

\usepackage{textcomp,marvosym}
\usepackage{booktabs}
\usepackage{cite}

\usepackage{nameref,hyperref}

\usepackage[right]{lineno}

\usepackage[nopatch=eqnum]{microtype}
\DisableLigatures[f]{encoding = *, family = * }

\usepackage[table]{xcolor}

\usepackage{array}

\newcolumntype{+}{!{\vrule width 2pt}}

\newlength\savedwidth

\raggedright
\usepackage[aboveskip=1pt,labelfont=bf,labelsep=period,justification=raggedright,singlelinecheck=off]{caption}

\makeatletter
\renewcommand{\@biblabel}[1]{\quad#1.}
\makeatother

\usepackage{lastpage,fancyhdr,graphicx}
\usepackage{epstopdf}
\fancyheadoffset[L]{2.25in}
\fancyfootoffset[L]{2.25in}
\begin{document}
\vspace*{0.2in}

\begin{flushleft}
{\Large
\textbf\newline{CBAM-SwinT-BL: Small Rail Surface Defect Detection Method Based on Swin Transformer with Block Level CBAM Enhancement} 
}
\newline
\\
Jiayi Zhao\textsuperscript{1},
Alison Wun-lam Yeung\textsuperscript{1},
Ali Muhammad\textsuperscript{1},
Songjiang Lai\textsuperscript{1},
Vincent To-Yee NG\textsuperscript{1*}

\bigskip
\textbf{1} Centre for Advances in Reliability and Safety(CAiRS), Hong Kong, China; 
\\
\bigskip

* Correspondence: vincent.ng@cairs.hk

\end{flushleft}
\section*{Abstract}
Under high-intensity rail operations, rail tracks endure considerable stresses resulting in various defects such as corrugation and spellings. Failure to effectively detect defects and provide maintenance in time would compromise service reliability and public safety. While advanced models have been developed in recent years, efficiently identifying small-scale rail defects has not yet been studied, especially for categories such as Dirt or Squat on rail surface. To address this challenge, this study utilizes Swin Transformer (SwinT) as baseline and incorporates the Convolutional Block Attention Module (CBAM) for enhancement. Our proposed method integrates CBAM successively within the swin transformer blocks, resulting in significant performance improvement in rail defect detection, particularly for categories with small instance sizes. The proposed framework is named CBAM-Enhanced Swin Transformer in Block Level (CBAM-SwinT-BL). Experiment and ablation study have proven the effectiveness of the framework. The proposed framework has a notable improvement in the accuracy of small size defects, such as dirt and dent categories in RIII dataset, with mAP-50 increasing by +23.0\% and +38.3\% respectively, and the squat category in MUET dataset also reaches +13.2\% higher than the original model. Compares to the original SwinT, CBAM-SwinT-BL increase overall precision around +5\% in the MUET dataset and +7\% in the RIII dataset, reaching 69.1\% and 88.1\% respectively. Meanwhile, the additional module CBAM merely extend the model training speed by an average of +0.04s/iteration, which is acceptable compared to the significant improvement in system performance.


\section*{Introduction}

Rail transportation serves as a critical element of contemporary transportation infrastructure. When trains operate for extended periods, the periodic contact and pressure between the train wheels and the rail surface can result in surface defects, such as corrugation, posing safety risks \cite{r1}. These defects are particularly prone to occur at stress-concentration points, such as phase joints, potentially resulting in track loosening and fractured rails. Consequently, the timely identification of defect sites through advanced technologies is imperative for enhancing operational efficiency, reducing labor costs, prolonging the lifespan of key components, and mitigating the risk of accidents \cite{r3}\cite{r4}. Traditional methods for the detection of rail surface defects predominantly rely on manual inspection, which produces inefficiency, high costs, and associated safety hazards. In recent years, various computer vision-based detection technologies have been developed to facilitate automated inspection of rail systems. Nonetheless, small-sized rail surface defects, such as dents and squats, present significant challenges for detection due to factors such as limited resolution, complex backgrounds, and surface noise. Thus, the identification of small defects in rail tracks remains a formidable challenge within the field. \\

To facilitate automated rail defect detection, various non-destructive testing technologies have been developed, which can be classified into five major approaches: geometric measurement, wheel track motion detection, changes in electromagnetic signals \cite{r5}\cite{r6}\cite{r8}, and computer vision-based detection methods. Compared to other detection techniques, computer vision-based approaches offer advantages in speed and efficiency while maintaining a completely non-invasive methodology. This advancement is made possible by improvements in camera monitoring technologies for railway systems, coupled with the development of sophisticated machine learning models, such as Convolutional Neural Networks (CNNs) and transformers.\\

Numerous studies have investigated CNN-based object detection models; however, they often face limitations in detecting small-sized defects. These CNN-based models can be classified into two categories based on their architectural structures: two-stage detectors and one-stage detectors \cite{r16}. The former, exemplified by the R-CNN series \cite{r17} and Feature Pyramid Networks \cite{r20}, provides higher accuracy and average recall rates for identifiable items. However, due to their complexity, these models are not commonly employed in real-time monitoring applications. In contrast, one-stage models, such as the YOLO series \cite{r21} and RetinaNet \cite{r22}, can detect all objects within a single inference step, thereby facilitating rapid responses and real-time detection \cite{r24}. Despite these advancements, both categories of CNN-based models encounter challenges in modeling global correlations due to their limited receptive fields, a significant drawback when detecting defects in small regions. \\

Recently, transformers, which have been primarily used in Natural Language Processing (NLP), have revolutionized the CV field and improved the detection of small-sized instances. These models utilize self-attention mechanisms that examine each pixel in relation to the entire image, allowing for the acquisition of global contextual information. This capability addresses certain limitations of Convolutional Neural Networks (CNNs) and improves model generalization.
The Vision Transformer (ViT) \cite{r25} leverages the transformer architecture as its backbone for image classification by segmenting the input image into fixed-size patches. These patches are then flattened into vectors and processed by the transformer. The hierarchical structure introduced by Pyramid ViT \cite{r26} and its subsequent iterations \cite{r27} has demonstrated effectiveness in dense prediction tasks such as object detection. Similarly, the Focal Transformer \cite{r28} and Cross Transformer \cite{r29} adopt a hierarchical architecture to capture both local and global visual signals.
Regarding local attention mechanisms, the Swin Transformer \cite{r30}\cite{r31} employs shifted windows to dynamically extract local information, contrasting with ViT's approach of partitioning the input image into uniformly sized patches. The shifted windows slide across the image, facilitating the capture of cross-window information essential for establishing global correlations. The Swin Transformer minimizes computational complexity at the attention layer through its shifted window methodology, achieving a state-of-the-art accuracy of 84.2\% on ImageNet, as reported.\\

Despite advancements in detection algorithms for small-sized targets, significant challenges persist under various conditions, such as those presented by aerial datasets. Several works have been proposed on the enhanced multiscale feature fusion method\cite{r62} and multi-layer feature concatenating\cite{r63}, or region proposed method that crops the object area and enlarges it to raw image size \cite{r64}. However, current work processes their experiment based on COCO dataset, where small area instances are defined by \(pixel < 32 \times 32\) and medium area as \(pixel < 96 \times 96\), with image resolution \(640 \times 480\). The equations are defined in equation \ref{eq1} and equation \ref{eq2}.\\

\begin{equation} \label{eq1}
 COCO\_small = 32 \times 32, relative\ size\ ratio = 0.3\%
\end{equation}

\begin{equation} \label{eq2}
 COCO\_medium = 96 \times 96, relative\ size\ ratio = 3\%
\end{equation}

The definition of small-size instance in COCO dataset is shown in formula. However, the same standard can not be employed directly on the rail surface datasets, where high resolution images are captured by high-speed cameras. In the following experiment, the definition of small size instance is the relative size \(< 2\%\), which means that Joint and Squat in MUET and all categories in RIII are considred as small size defects.\\

Various solutions have been proposed based on RCNN and YOLO \cite{r32}\cite{r33}, as well as additional modules focusing on features such as feature fusion \cite{r34} and feature texture transfer \cite{r35}. Moreover, research has explored the integration of attention mechanisms within CNN-based models. The Convolutional Block Attention Module (CBAM) \cite{r36} has been developed as a lightweight and efficient module that can be incorporated into any CNN architecture to enhance the extraction of spatial and channel information. It has been proposed as an augmentative feature in YOLOv5, allowing for the direct training of the module alongside the base network, thereby producing higher-resolution feature maps for small objects \cite{r37}. However, the YOLO series, being CNN-based detectors, face challenges in capturing global features. \\

In contrast, the CBAM module has demonstrated effectiveness when integrated with the Swin Transformer in image segmentation tasks, for instance, within the UNet framework for medical CT images \cite{r38}. Additionally, CBAM has been utilized at the Neck of a Swin Transformer-based architecture for high-accuracy ship detection in synthetic aperture radar images \cite{r39} and combined with the Swin Transformer prior to patch partitioning for emotion classification tasks \cite{r40}. Notably, deploying CBAM immediately after the input image or features, as has been done in most studies, may not yield optimal results. This limitation arises because the CBAM module, designed on traditional CNN frameworks, may have reduced capability in handling high-resolution images. Conversely, integrating CBAM at the patch merging stage of the Swin Transformer—such as for building segmentation in remote sensing images \cite{r41}—could enhance performance. Nonetheless, the attention information extracted by CBAM might diminish after passing through the MLP layer of the Window Multi-Head Self Attention block, potentially affecting its efficacy during the computation of the Shifted Window Multi-Head Self Attention. Furthermore, the combination of CBAM and the Swin Transformer, particularly at the backbone level, remains underexplored in the current literature. Thus, further research is needed to investigate the potential advantages of a CBAM-enhanced Swin Transformer and its integration methods.\\

Alongside enhancing object detection models, the quality of rail track datasets is critical. Many railway surface datasets possess a limited number of images \cite{r42}. For example, the CRRC dataset \cite{r44} includes approximately 1,000 images, while the Rail-5k dataset \cite{r45} also features around 1,000 images across 13 classes. Consequently, this study utilizes two public rail surface defect datasets provided by the NCRA Condition Monitoring Systems Lab at Mehran University of Engineering and Technology (MUET) \cite{r54} and the Railway Infrastructure Inspection Institute (RIII) \cite{r55}, which offer a greater number of higher-resolution and more recent images. Nonetheless, effective data preprocessing remains essential to further address these limitations. \\

Regarding current object-detection-based approaches for rail defect detection, several techniques are evident in the literature. One effective strategy involves pretraining on large-scale foundational datasets, such as MS COCO \cite{r46} and Imagenet \cite{r47}, followed by fine-tuning to meet specific operational requirements \cite{r48}. Alternatively, employing a segmentation decoder with knowledge transfer can enhance detection accuracy in crack detection scenarios \cite{r49}. A transformer-based method utilizing multiple backbones to capture distinct features has been proposed for multi-class detection, while other studies have fine-tuned transformers to produce five output scales, making them more suitable for rail defect detection with significant target size variations \cite{r50}. TSSTNet introduces a two-stream encoder for improved feature and edge extraction, facilitating better defect positioning and contouring \cite{r51}. Lightweight models based on YOLO have also been considered for real-time defect detection. The BiFPN module, integrated into YOLOX with the attention mechanism NAM, has been evaluated on the RSDDs dataset \cite{r52}. Additionally, improvements to YOLOv7-tiny with BiFPN and ECA attention mechanisms have been assessed on the GC10-DET and NEU-DET datasets \cite{r53}. This paper employed CBAM-enhanced Swin-T model for rail defect detection, utilizing data augmentation as well as image enhancement techniques for data preprocessing, and integrate attention modules separately in block level to achieve better detection performance on small size categories, such as Dirt and Squat on rail surface.\\

The paper is organized in the following steps: Section two describes the methodology of Channel Attention Module and Spatial Attention module in CBAM and the architecture of Swin Transformer model, where the necessity as well as feasibility shall be discussed. Section three give a statistic analysis and a visualization on public MUET and RIII dataset. Data augmentation and image enhancement method were introduced for data preprocessing. Section four present model parameter setting and performance evaluation in ordinary swin transformer as well as CBAM enhanced in Model, Stage and Block Level. The visualization images of the model output were carried out. Section five is the conclusion of the paper.\\



\section*{Methodology}
\subsection*{Overview}

To address the aforementioned limitations, this research integrate CBAM and Swin Transformer in multiple functions and found that CBAM enhancement in Block Level preform better detection result in small size instances, which has also been proved by experiment on open source rail surface dataset MUET and RIII.\\

\subsection*{Convolutional Block Attention Module}
The Convolutional Block Attention Module (CBAM) is a novel attention mechanism that improve neural network’s performance in visual recognition tests. While traditional CNN has shown impressive performance in object detection and image classification, it has trouble capturing fine grained spatial details and effectively highlight informative channels within feature maps, and a potential remedy for these issues is the development of attention mechanisms.\\

Complementary attention modules contribute to CBAM for the function of detecting small-size items. The Channel Attention Module (CAM) obtains channel-specific attention by adaptively re-calibrating feature maps based on the relevance of each channel. It employs parallelized pooling layers followed by fully connected layers to model channel dependencies. In contrast, Spatial Attention Module (SAM) grabs spatial attention by learning to prioritize informative spatial places. It utilizes max pooling and average pooling operations to capture both maximum and average responses within feature maps.\\

\begin{figure}
    \centering
    \includegraphics[width=1\linewidth]{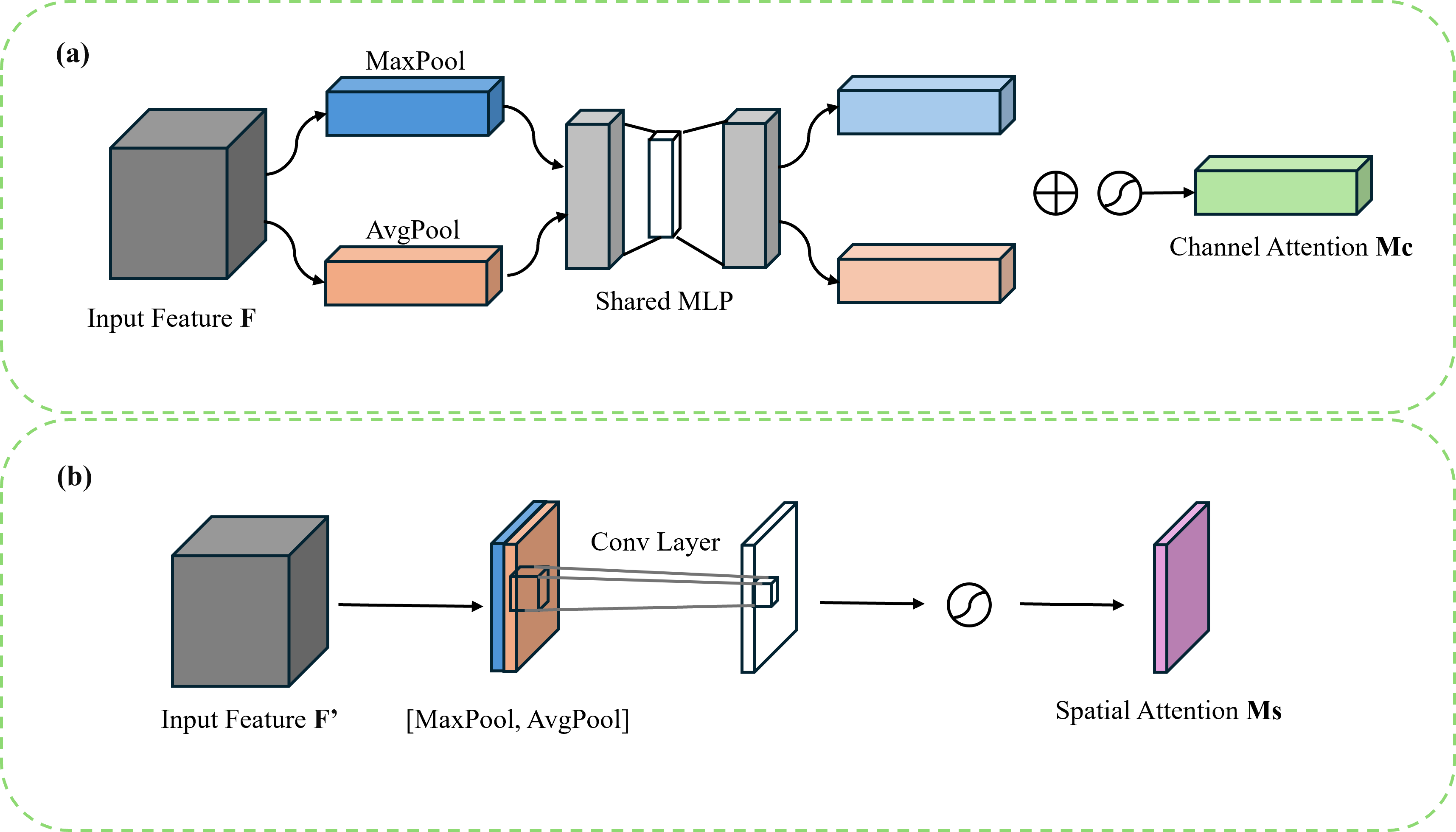}
    \caption{{\bf CBAM module architecture.} (a) Channel attention module. (b) Spatial attention module.}
    \label{fig:CBAM module architecture}
\end{figure}

CBAM framework is shown in Figure 1, where the feature \(F\) with vector shape \(F \in \mathbb{R}^{C\times H \times W}\) was denoted as input. In channel attention, the average-pooling and max-pooling layers are organized in parallel, aggregating the spatial information of the feature map and producing two context descriptors, \(F_{avg}^{c}\) and \(F_{max}^{c}\), which indicate the average and maximum pooling features, respectively. The spatial information of the feature map can also be generated through two pooling layers, \(F_{avg}^{s}\) and \(F_{max}^{s}\), which represent the average pooling feature and maximum pooling feature of each channel. The calculated vectors shall be fed into a Multi Layer Perception (MLP) with shared parameters, or a Convolutional Network depends on either sub module, then merged into output block using element-wise summation. The channel attention mechanism and spatial attention mechanism can be computed in equation \ref{eq3} and equation \ref{eq4}:\\

\begin{equation} \label{eq3}
\begin{split}
M_c(F)
& = \delta ( MLP(AvgPool(F)) + MLP(MaxPool(F)))\\ 
& = \delta(W_1(W_0(F_{Avg}^{c})) + W_1(W_0(F_{Max}^{c})))
\end{split}
\end{equation}

\begin{equation} \label{eq4}
\begin{split}
M_s(F)
& = \delta (\mathnormal{f}^{7 \times 7}[AvgPool(F);MaxPool(F)])\\ 
& = \delta(\mathnormal{f}^{7 \times 7}([F_{Avg}^{s}; F_{Avg}^{s}]))
\end{split}
\end{equation}

The output of sub-model was denoted as  \(M_{c} \in \mathbb{R}^{C\times 1 \times 1}\) and  \(M_{s} \in \mathbb{R}^{1\times H \times W}\) subsequently. \(\delta\) denotes the sigmoid function in the formula. \(W_0\) and \(W_1\)  are the weight of two MLP layers and \(\mathnormal{f}^{7 \times 7}\) represent a convolution operation with filter size \(7 \times 7\).\\

\subsection*{Swin Transformer}

The Swin Transformer architecture is built based on transformer module, which received breakthrough achievements on natural language processing back to early 1990s, was extended to computer vision projects such as object detection and image classification. It eliminates the limitations of traditional Convolutional neural networks by leveraging self-attention mechanisms, which efficiently capture global contextual information for model training.\\

The shifted window mechanism is incorporated in Swin Transformer, which allows images to be divided into several non-overlapping patches with small size, and the attention head shift between patches to achieve higher detection performance, as opposed to the traditional method of processing the whole image at once. With its hierarchical structure and shifted attention-based methodology, swin transformer reach the cutting-edge performance on various computer vision tasks, such as semantic segmentation, object detection and image classification. It has demonstrated exceptional scalability and robustness make it a promising architecture for advancing the field of computer vision.\\

\begin{figure}
    \centering
    \includegraphics[width=1\linewidth]{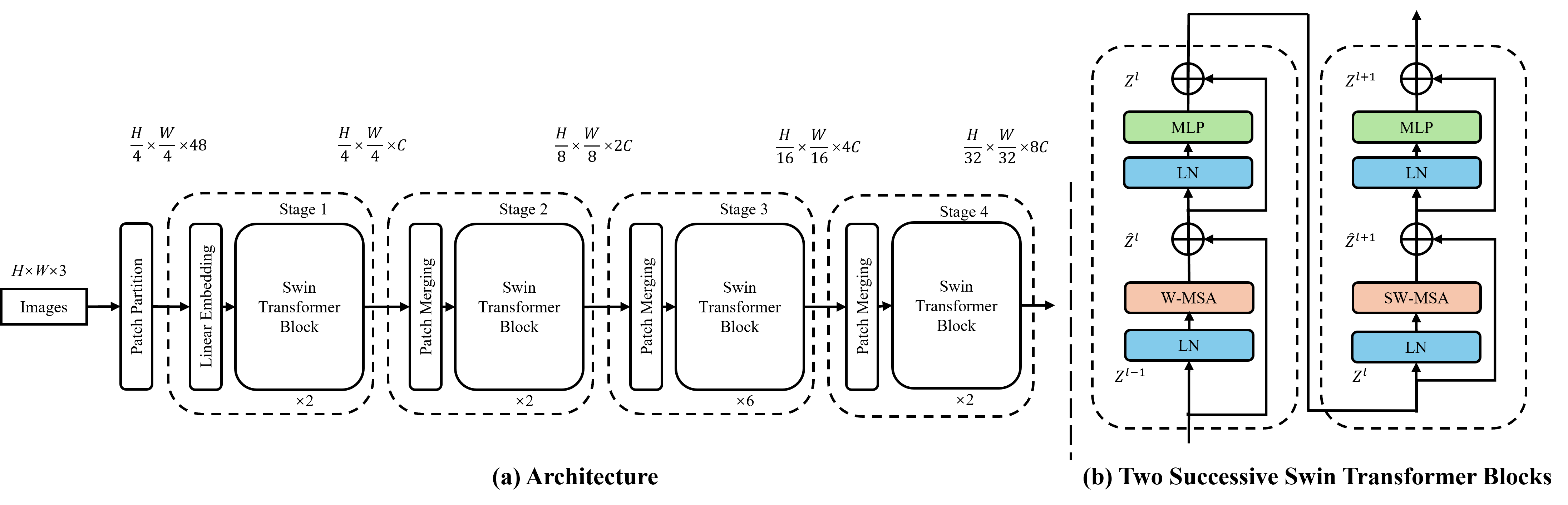}
    \caption{{\bf The architecture of Swin Transformer model.}
    (a) General structure of Swin Transformer. (b) two successive Swin Transformer blocks.}
    \label{fig:Swin Transformer}
\end{figure}

Figure 2 shows the structure of a standard Swin Transformer\cite{r30}, using Layer Normalization (LN) followed by regular window based Multi-Head Attention (W-MSA) and Shifted Window Multi-Head Attention (SW-MSA). The general framework was made up of four stages connected in series, with each stage separated into two parts: Patch Merging and successive Swin Transformer blocks. In patch merging section, patch partition splits the input vector four times in the spatial directions and subsequently flattened in the channel direction. Then a feature map with a size of \(\frac{H}{4} \times \frac{H}{4} \times \mathnormal{C}\) will be generated by the linear embedding layer, utilizing \(1 \times 1\) convolution to convert the channel dimension to \(\mathnormal{C}\). The continuous Swin Transformer Blocks can be computed in equation \ref{eqn-5}:\\

\begin{align}
      & \widehat{z}^{l} = W - MSA\Bigl(LN(z^{l-1}) \Bigr) + z^{l-1} \nonumber\\
      & z^{l} = MLP\Bigl(LN(\widehat{z}^{l}) \Bigr) + \widehat{z}^{l} \nonumber\\
      & \widehat{z}^{l+1} = SW - MSA\Bigl(LN(z^{l}) \Bigr) + z^{l}\nonumber\\
      & z^{l+1} = MLP\Bigl(LN(\widehat{z}^{l+1}) \Bigr) + \widehat{z}^{l+1} \label{eqn-5}
\end{align}

The output feature of Multi-head Self-Attention (W-MSA/SW-MSA) is defined as \(\widehat{z}^{l}\) and the output of Multi-Layer Perception (MLP) at block \(l\) is defined as \(z^{l}\). LN denoted as Layer Normalization. Up-sampling and down-sampling techniques are generally employed to enlarge or reduce the size of input image, to improve the fusion of feature maps on different image sizes as well as ensuring the resolution remains constant. Semantic information loss and local interference are prevalent problems caused by feature map compression, which has become increasingly significant in small-scale object detection tasks. Considering an approach to improve object detection performance, the CBAM module focuses on deep level features and increases the system robustness by incorporating a combination of average pooling and maximum pooling on channel and spatial, respectively.

\subsection*{CBAM enhanced Swin Transformer in deep level}
The attention-block-enhanced approaches for object detection have been applied before, while the combination method between CBAM and Swin Transformer block has not yet produced definitive results. Given the properties of convolutional neural networks, the efficiency of feature extraction on large-scale input image is lower than attention block. As a result, CBAM may not be able to operate its fully potential by being simply assigned on the input vector side.\\

\begin{figure}
    \centering
    \includegraphics[width=1\linewidth]{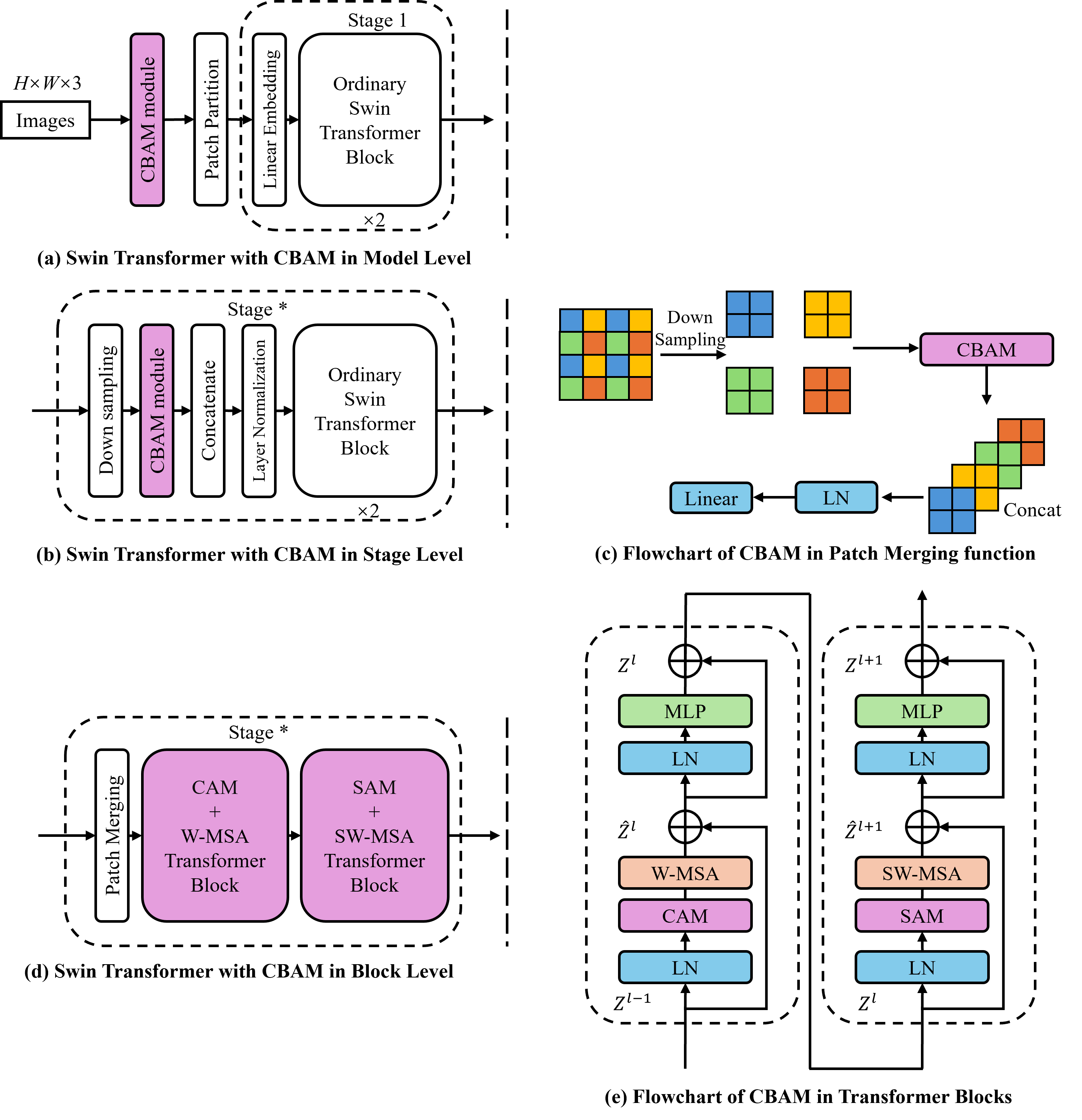}
    \caption{{\bf CBAM enhanced Swin Transformer framework.} (a) CBAM in Model Level. (b) CBAM in Stage Level. (c) Flowchart of Patch Merging in Stage Level. (d) CBAM in Block Level. (e) Flowchart of CAM and SAM sequentially located inside Swin Transformer Blocks.}
    \label{fig:Swin Transformer with CBAM}
\end{figure}

Figure 3 (a) display the structure of model-level optimization, where the module is simply proposed between the input image and the Swin Transformer. The stage-level structural is represented in Figure 3 (b) (c), where CBAM module is inserted between patch partition and linear embedding to optimize the divided input vector before concatenating and normalization. The two approaches mentioned above have been frequently used to integrate CBAM with detection models, well the CBAM module can also be operated directly on the Swin Transformer Block. To improve responses under static detection circumstances, it is recommended to establish a better balance between the computation cost on model training and the detection quality for small size objects. The recommendation offering is to optimize the structure at the block level as shown in Figure 3 (d) (e). W-MSA and SW-MSA are positioned after CAM and SAM respectively, where the concatenation and normalization vector set instead of the original input image will be properly operated upon the two modules. This will make more efficient of SAM's capability to magnify changed pixels in the spatial dimension and CAM's capacity to promote high-value channels from limit insignificant ones.\\

\begin{table}[]
\begin{adjustwidth}{-2.0in}{0in}
\centering
{%
\begin{tabular}{|lll}
\hline
\multicolumn{3}{|c|}{\textbf{\begin{tabular}[c]{@{}l@{}}Algorithm 1: CBAM enhanced Swin Transformer in Block Level\end{tabular}}}                \\ \hline
   & \multicolumn{1}{l|}{\textbf{\begin{tabular}[c]{@{}l@{}}Input: Patch embedded vector x; Shape of the \\ vector x.shape = b, h, w, c;\end{tabular}}}    & \multicolumn{1}{l|}{}     \\
   & \multicolumn{1}{l|}{\begin{tabular}[c]{@{}l@{}}Output: SwinBlockSequence(depth = 2); \\ \quad \# 2 Successive block as an example\end{tabular}} & \multicolumn{1}{l|}{}\\
1  & \multicolumn{1}{l|}{\textbf{class ChannelAttention(BaseModule):}} & \multicolumn{1}{l|}{}                                                       \\
2  & \multicolumn{1}{l|}{\quad def forward(self, channels):}  & \multicolumn{1}{l|}{}                                                                \\
3  & \multicolumn{1}{l|}{\qquad \textit{… \# Produce channel-wise attention map}}  & \multicolumn{1}{l|}{}                                           \\
4  & \multicolumn{1}{l|}{\quad \textbf{return} output}   & \multicolumn{1}{l|}{}                                                                     \\
5  & \multicolumn{1}{l|}{\textbf{class SpatialAttention(BaseModule):}}    & \multicolumn{1}{l|}{}                                                    \\
6  & \multicolumn{1}{l|}{\quad def forward(self, channels):}           & \multicolumn{1}{l|}{}                                                       \\
7  & \multicolumn{1}{l|}{\qquad \textit{… \# Produce spatial-wise attention map}}   & \multicolumn{1}{l|}{}                                          \\
8  & \multicolumn{1}{l|}{\quad \textbf{return} output}        & \multicolumn{1}{l|}{}                                                                \\
9  & \multicolumn{1}{l|}{\textbf{class SwinBlockSequence (BaseModule):}}       & \multicolumn{1}{l|}{}                                               \\
10 & \multicolumn{1}{l|}{\quad \textbf{def \_\_init\_\_(self,):}}            & \multicolumn{1}{l|}{}                                                 \\
11 & \multicolumn{1}{l|}{\qquad self.channel\_attention = ChannelAttention()}     & \multicolumn{1}{l|}{}                                            \\
12 & \multicolumn{1}{l|}{\qquad self.spatial\_attention = SpatialAttention()}  & \multicolumn{1}{l|}{}                                               \\
13 & \multicolumn{1}{l|}{\quad \textbf{def forward(self, x, hw\_shape):}}       & \multicolumn{1}{l|}{}                                              \\
14 & \multicolumn{1}{l|}{\qquad shortcut = x}            & \multicolumn{1}{l|}{}                                                                     \\
15 & \multicolumn{1}{l|}{\qquad x = self.norm1(x) } & \multicolumn{1}{l|}{\textit{\#Layer normalization}}              \\
16 & \multicolumn{1}{l|}{\qquad x = self.channel\_attention(x, channels = C) }    &  \multicolumn{1}{l|}{\textit{\#Compute CAM}}                    \\
17 & \multicolumn{1}{l|}{\qquad x = self.w\_msa(x\_windows , mask = None)  }      &   \multicolumn{1}{l|}{\textit{\#Compute W-MSA}}         \\
18 & \multicolumn{1}{l|}{\qquad x = x + shortcut \#Residual network}   & \multicolumn{1}{l|}{}                                              \\
19 & \multicolumn{1}{l|}{\qquad x = x + self.mlp(self.norm2(x)) }   & \multicolumn{1}{l|}{\textit{\#Layer normalization, MLP, and residual network}} \\
20 & \multicolumn{1}{l|}{\quad \textit{\# End of the first block}}     & \multicolumn{1}{l|}{}                                                       \\
21 & \multicolumn{1}{l|}{\qquad shortcut = x}               & \multicolumn{1}{l|}{}                                                                  \\
22 & \multicolumn{1}{l|}{\qquad x = self.norm1(x)  }         & \multicolumn{1}{l|}{\textit{\#Layer normalization}}                  \\
23 & \multicolumn{1}{l|}{\qquad x = self. spatial\_attention(x, channels = 2) }  & \multicolumn{1}{l|}{\textit{\#Compute SAM}}                     \\
24 & \multicolumn{1}{l|}{\qquad x = self.sw\_msa(x\_windows, mask = attn\_mask) }      &\multicolumn{1}{l|}{\textit{\#Compute SW-MSA}}            \\
25 & \multicolumn{1}{l|}{\qquad x = x + shortcut } &    \multicolumn{1}{l|}{\textit{\#Residual network}}                                           \\
26 & \multicolumn{1}{l|}{\qquad x = x + self.mlp(self.norm2(x)) } & \multicolumn{1}{l|}{\textit{\#Layer normalization, MLP, and residual network}}   \\
27 & \multicolumn{1}{l|}{\quad \textit{\# End of the second block}}     & \multicolumn{1}{l|}{}                                                      \\
28 & \multicolumn{1}{l|}{\textbf{return x;}}                        & \multicolumn{1}{l|}{}                                \\ \hline
\end{tabular}
}
\end{adjustwidth}
\end{table}

The pseudocode of CBAM enhanced Swin Transformer in Block level, as shown in Fig. 3(d), is described in Algorithm 1. The input feature x is the small-scale vector after patch partition and linear embedding, and \(h and w\) are the vector size. Parameter \(c\) is the correlation coefficient of the embedding dimension.\\

The input vector x is initially normalized before being transmitted through the channel attention module and spatial attention module in accordance with the schedule as the swin transformer model is configured. Subsequently, a Multi-head Self-Attention computation is performed by a standard transformer approach, followed by data integration and fully connected layer. The output from the algorithm loop will be evaluated as the model's result when execution is finished, and the aforementioned procedure shall be repeated until model reach convergence.\\

\section*{Experiment}
\subsection*{Rail Surface Defect Dataset}

Advanced Rail Transportation system has established image capturing system on the railway transit network, in order to monitor the operation condition of the rail track. The traditional approach relies on manual inspection of recorded images, which is time-consuming and requires a large number of front-line operators. \\

The labelled image data was collected from two open-source dataset, namely the MUET and the RIII dataset. MUET is provided by the NCRA Condition Monitoring Systems Lab at Mehran University of Engineering and Technology from Pakistan in 2024. The dataset is acquired using an EKENH9R camera, with the inspection vehicle's maximum speed limited to 20 km/h. It helps guarantee that the image is crisp while ensuring every shot has an observable surface defect. It contains 5157 RGB outdoor rail images with a resolution of 1280 \(\times\) 720. The defects are divided into 7 categories: Crack, Flaking, Groove, Joint, Shelling, Spalling, and Squat, as shown in Figure 4. \\

RIII is provided by the Railway Infrastructure Inspection Institute from Mainland China. The images in the dataset, shown in Figure 5, was captured by a linear array camera aboard a high-speed train on a 9-kilometer test route. 400 grayscale tunnel rail images with defect features with a size of 2048 \(\times\) 2000 were labeled in the dataset. The dataset contains 8 recognized defects: Damage, Dirt, Gap, Dent, Crush, Scratch, Slant, and unknown. Unknown indicates that it is a highly suspected defect, but its type cannot be determined, requiring physical verification. It should be emphasized that the RIII dataset need to be augmented before model training to avoid over fitting in limit amount of data, which may also lead to better system performance. \\

\begin{figure}[!h]
\includegraphics[width=1\linewidth]{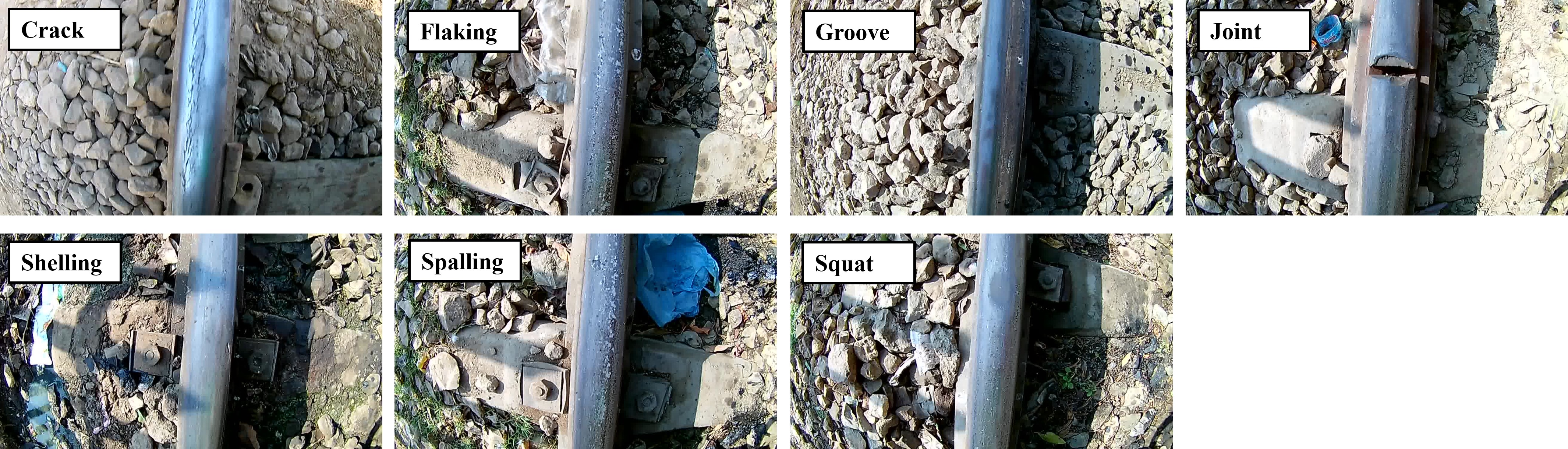}
\caption{{\bf Samples of the MUET dataset with 7 categories.}
From left to right, the first row contains Crack, Flaking, Groove and Joint, and the second row contains Shelling, Spalling and Squat.}
\label{fig:MUET_sample}
\end{figure}

\begin{figure}[!h]
    \centering
    \includegraphics[width=1\linewidth]{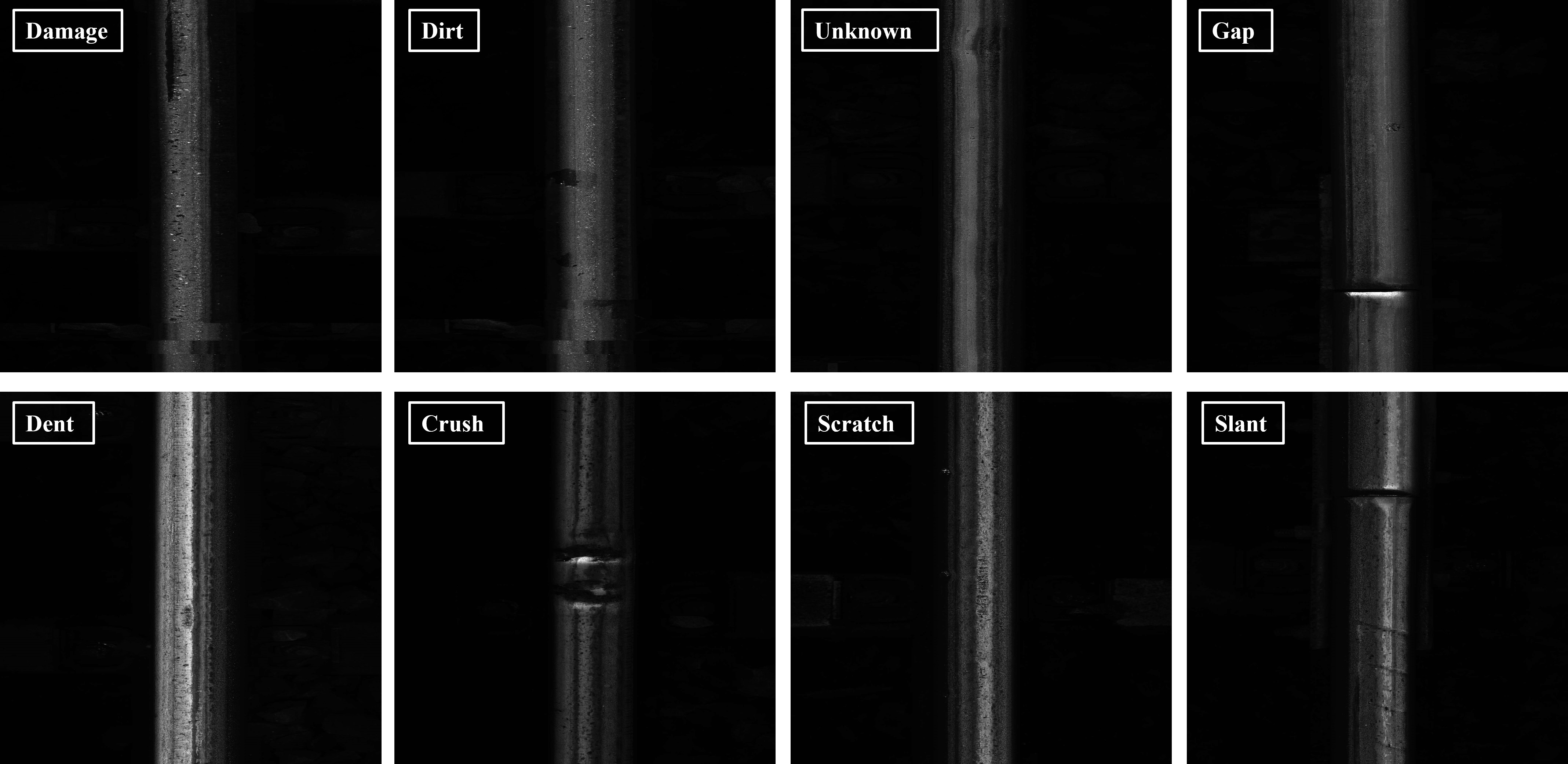}
    \caption{{\bf Samples of the RIII dataset with 8 categories.}
    From left to right, the first row contains Damage, Dirt, Unknown and Gap, and the second row contains Dent, Crush, Scratch and Slant.}
    \label{fig:RIII_sample}
\end{figure}

Similar to traditional object detection dataset datasets, rail defect images are annotated to identify a broad range of valuable rail system components as well as potential rail track defects. Statistical analysis demonstrates that the number of images between categories is not balanced. Considering that a single image may contain multiple instances, the number of instances for each category is recorded in Figure 6 and Figure 7. \\

\begin{figure}
	\begin{minipage}[t]{1\linewidth}
		\centering
		\includegraphics[width=5.5in]{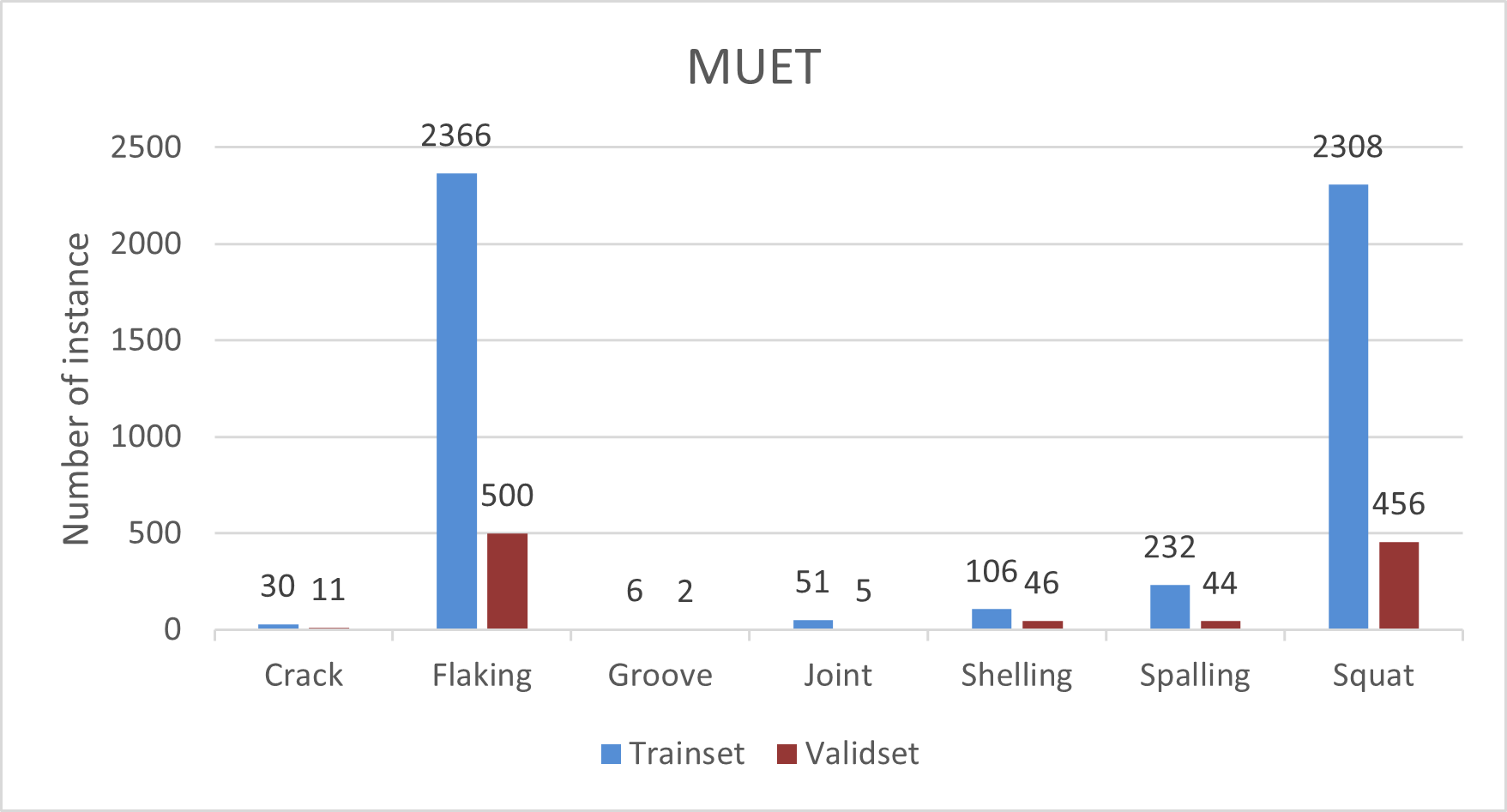}
		\caption{{\bf Number of instances in MUET Dataset\\}}
		\label{fig:MUET_distribution}
	\end{minipage}
	\\
	\begin{minipage}[t]{1\linewidth}
		\centering
		\includegraphics[width=5.5in]{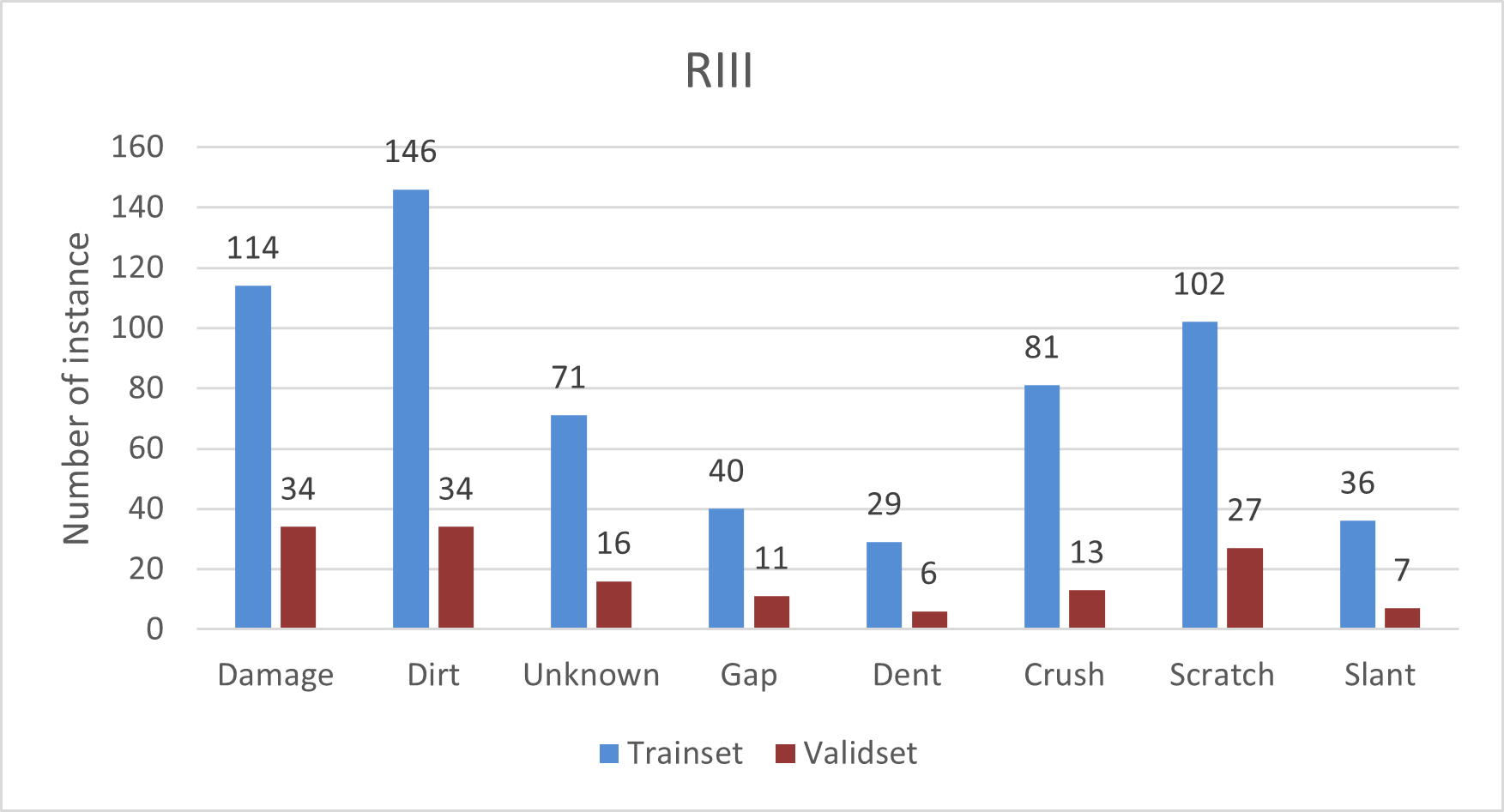}
		\caption{{\bf Number of instances in RIII Dataset}}
		\label{fig:RIII_distribution}
	\end{minipage}
\end{figure}

It should be noted that each defect type has a distinct average size. For example, the judgment area of the category “Shelling” at the surface of the rail track in the MUET dataset is relatively large, while the “Squat” resulting from the long-term friction between wheels and rail track has a small judgment area. Theoretically, entities with smaller sizes are more difficult to be identified. The definition of labelled categories with instance numbers is given in Table 1 and Table 2. Avg\_Size by Pixel is the absolute size of the bounding box per category, and the Avg\_Size by Ratio is the relative size compared to the whole image. The definition is shown in equation \ref{eq6} and equation \ref{eq7}: \\



\begin{eqnarray}
\label{eq6}
	Avg\_Size\ by\ Pixel = \frac{1}{N}\sum_0^N w_{ij} \times h_{ij}
\end{eqnarray}

\begin{eqnarray}
\label{eq7}
	Avg\_Size\ by\ Ratio = \frac{1}{N}\sum_0^N \frac{w_{ij} \times h_{ij}}{W_i \times H_i}
\end{eqnarray}

The symbol \(w_{ij}\) and \(h_{ij}\) describe the width and height of j-th bounding box in i-th image. and \(W_{ij}\) and \(H_{ij}\) denote the width and height of image i.

\begin{table}[]
\begin{adjustwidth}{-2.25in}{0in}
\centering
\caption{{\bf Category definition and data analysis of MUET Dataset.}}
\label{tab:MUET Dataset}
\resizebox{1.4\columnwidth}{!}
{%
\begin{tabular}{cccccc}
\hline
\textbf{Dataset - MUET} & \textbf{Attr.} & \textbf{Definition}                                          & \textbf{\begin{tabular}[c]{@{}c@{}}Num. of \\ instance\end{tabular}} & \textbf{\begin{tabular}[c]{@{}c@{}}Avg\_Size \\ by Pixel\end{tabular}} & \textbf{\begin{tabular}[c]{@{}c@{}}Avg\_Size \\ by Ratio\end{tabular}} \\ \hline
\textbf{Crack}          & Defect         & Break on the rail surface caused by wheel contact.           & 41                                                                   & 87 × 634                                                               & 6.0\%                                                                  \\
\textbf{Flaking}        & Defect         & Rail surface material fall away in flakes                    & 2866                                                                 & 132 × 690                                                              & 9.9\%                                                                  \\
\textbf{Groove}         & Defect         & Narrow cut with vertical depth in rail track                 & 8                                                                    & 107 × 714                                                              & 8.3\%                                                                  \\
\textbf{Joint}          & Normal         & Joint of two rail track sections without welding             & 56                                                                   & 172 × 81                                                               & 1.5\%                                                                  \\
\textbf{Shelling}       & Defect         & Small/mild metal remove on rail track surface                & 152                                                                  & 141 × 680                                                              & 10.4\%                                                                 \\
\textbf{Spalling}       & Defect         & Big/severe metal remove on rail track surface                & 276                                                                  & 145 × 708                                                              & 11.2\%                                                                 \\
\textbf{Squat}          & Defect         & Surface defect caused by metal fatigue during wheel contact. & 2764                                                                 & 96 × 125                                                               & 1.3\%                                                                  \\ \hline
\end{tabular}%
}
\end{adjustwidth}
\end{table}

\begin{table}[]
\begin{adjustwidth}{-2.25in}{0in}
\centering
\caption{{\bf Category definition and data analysis of RIII Dataset.}}
\label{tab:RIII Dataset}
{%
\begin{tabular}{cccccc}
\hline
\textbf{Dataset - RIII} & \textbf{Attr.} & \textbf{Definition}                                   & \textbf{\begin{tabular}[c]{@{}c@{}}Num. of \\ instance\end{tabular}} & \textbf{\begin{tabular}[c]{@{}c@{}}Avg\_Size \\ by Pixel\end{tabular}} & \textbf{\begin{tabular}[c]{@{}c@{}}Avg\_Size \\ by Ratio\end{tabular}} \\ \hline
\textbf{Damage}         & Defect         & Tear of the lateral planes of the railhead            & 148                                                                  & 137 × 116                                                              & 0.39\%                                                                 \\
\textbf{Dirt}           & Normal         & Paint, or mud that covers the surface of the rail     & 180                                                                  & 162 × 181                                                              & 0.72\%                                                                 \\
\textbf{Unknown}        & Normal         & Unrecognized features                                 & 87                                                                   & 120 × 93                                                               & 0.27\%                                                                 \\
\textbf{Gap}            & Normal         & Gaps left between successive rails on a railway track & 51                                                                   & 416 × 83                                                               & 0.84\%                                                                 \\
\textbf{Dent}           & Defect         & Defect initiated from rolling contact fatigue cracks  & 35                                                                   & 130 × 142                                                              & 0.45\%                                                                 \\
\textbf{Crush}          & Defect         & Big/severe wear of the lateral planes of the railhead & 94                                                                   & 415 × 126                                                              & 1.28\%                                                                 \\
\textbf{Scratch}        & Defect         & Small/mild wear of the lateral planes of the railhead & 129                                                                  & 98 × 404                                                               & 0.97\%                                                                 \\
\textbf{Slant}          & Defect         & Displacement of the parent metal from the railhead    & 43                                                                   & 257 × 87                                                               & 0.55\%                                                                 \\ \hline
\end{tabular}%
}
\end{adjustwidth}
\end{table}

\subsection*{Data Preprocessing}
As the railway system designed to travel through tunnels, collecting rail surface images in low-light conditions might result in low contrast and indistinct boundaries between detected features and the background. As a result, it is necessary to improve the quality of rail surface images through technological methods, as well as use data enhancement algorithms to perform image preprocessing on the original images, which can highlight the distinctive information of rail defects and effectively improve the defect detection rate for object detection. The image enhancement algorithm is shown in the Figure 8. \\

\begin{figure}
    \begin{adjustwidth}{-2.25in}{0in}
	\begin{minipage}[t]{1\linewidth}
		\centering
		\includegraphics[width=7.5in]{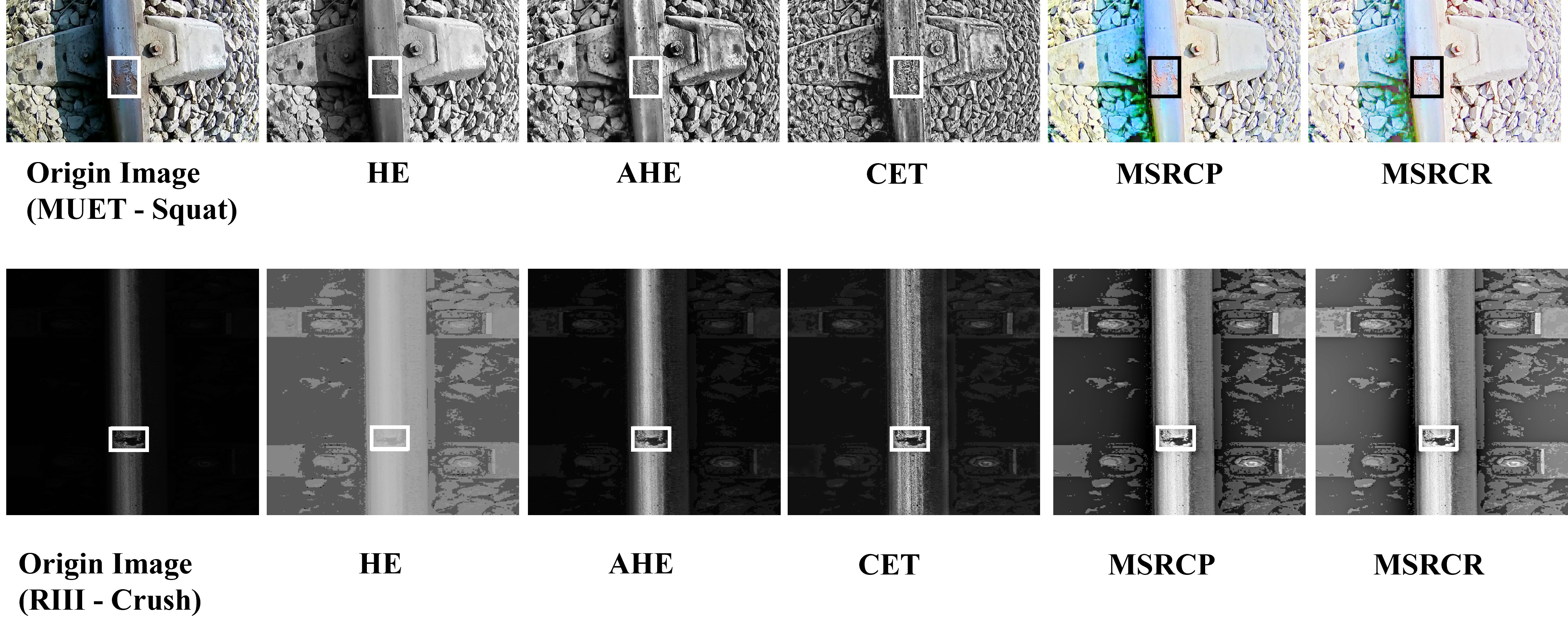}
		\caption{{\bf Visualization of Data Enhancement\\}}
		\label{fig:Data_enhancement}
	\end{minipage}
	\\
	\begin{minipage}[t]{1\linewidth}
		\centering
		\includegraphics[width=7.5in]{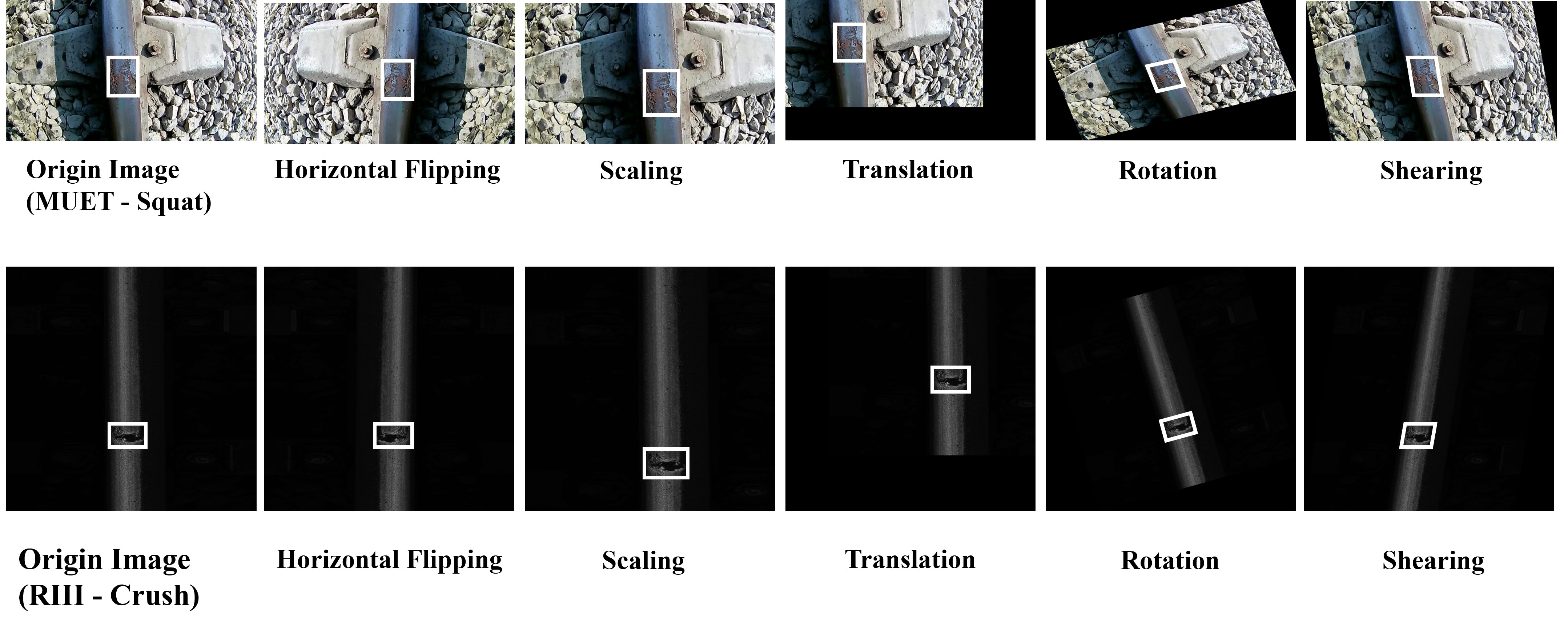}
		\caption{{\bf Visualization of Data Augmentation}}
		\label{fig:Data_augmentation}
	\end{minipage}
    \end{adjustwidth}
\end{figure}

The image above provides a visual representation of data enhancement for images with the label Squat and Crush, with an expanded view of the effective regions highlighted as objects. It shows that the data enhancement method increases the recognition of essential components. Histogram Equalization (HE) and Adaptive Histogram Equalization (AHE) are the common image enhancement methods; CET\cite{r56} denotes Contrast Enhancement Techniques; MSRCP and MSRCR\cite{r57} represent Multi-Scale Retinex with Color Preservation and Multi-Scale Retinex with Color Restoration. Given that the training images in the RIII dataset are in grayscale, the data enhancement method can be chosen without considering the supplementary colour space approach. As a result, enhancement method MSRCP is used for MUET dataset and CET for RIII dataset. \\

The variation in the quantity of training data between categories is an important consideration influencing the overall system performance in multi-category object detection tasks. Smaller data sets often suffer from overfitting because of the model's tendency to fit the category with more data during the continuous training process. Data augmentation has become an essential approach for enhancing model performance on object detection with limited training data. Ordinary yet efficient modifications, such as flipping, scaling and rotation, are used in the experiment considering the environment background of the rail image dataset. As shown in Figure 9, a random mix of several improvement approaches has been employed to increase training data diversity. \\

Rail surface datasets are collected based on real operating environment, variations in the quantity of regular railway components and rail surface defects are common occurrences in railway scenarios. Consequently, for categories with limited data, data augmentation shall be processed as a standard approach. The target is to have at least 400 images in training set and 100 images in validation set after train-test split. Table 3 shows the information of image data after data preprocessing. \\

\begin{table}[]
\centering
\caption{{\bf Number of images in rail defect datasets after data augmentation.}}
\label{tab:data augmentation}
\begin{tabular}{ccc}
\hline
\textbf{Dataset}               & \textbf{MUET} & \textbf{RIII} \\ \hline
\textbf{Trainset}              & 4186          & 320           \\
\textbf{Trainset augmentation} & 5836          & 2211          \\
\textbf{Validset}              & 967           & 80            \\
\textbf{Validset augmentation} & 1456          & 641           \\ \hline
\end{tabular}
\end{table}

\subsection*{Experiment Setting}

Rail surface image data are enhanced, augmented (scaling, translation, rotation, and shearing), and transformed into COCO format for model training. For this task of rail defect detection, Mask-RCNN was used as the detector, with CBAM-enhanced Swin Transformer (Swin-T) as the backbone for feature extraction. The proposed method, together with other baseline models, such as original Mask-RCNN and YOLO are evaluated based on the 80-20 separated training set. The optimizer transformer employed is AdamW with a learning rate of 0.0001, with a corresponding weight decay of 0.05. Given the limited amount of rail images, the training epoch is set to 36 to avoid overfitting. Pre-training method was performed on COCO dataset before introducing rail surface dataset for better model performance. Swin transformer the tiny version is used in the experiment, considering the scale of training dataset. \\

\subsection*{Model Performance and Ablation Study}

The evaluation metrics of object detection are shown in equation \ref{eqn-8}, where the \(TP = True\ Positive\), \(TN = True\ Negative\), \(FP = False\ Positive\) and \(FN = False\ Negative\). \(mAP\) and \(mAR\) can be computed as the sum of \(AP/AR\) in each category \(i\) divided by the number of categories \(N\). Precision evaluate the accuracy of the model's positive predictions and recall indicates the ability to identify all related instances. To be more specific, Mean Average Precision(mAP) offering a more comprehensive evaluation in different detection scenarios, by computing the average AP across multiple confidence thresholds instead of a fixed value. The \(mAP.50\) in the following experiment calculate mAP of each model with a Intersection over Union(IoU) overlap of at least 0.50(50\%) and \(mAP.75\) requiring a minimum IoU overlap of 0.75(75\%), which evaluates the model's ability to localize instances with higher precision.\\

\begin{align}
      & Precision = \frac{TP}{TP + FP} \nonumber\\
      & Recall = \frac{TP}{TP + FN} \nonumber\\
      & mAP = \frac{1}{N}\sum_{i=1}^{N} AP_{i} \nonumber\\
      & mAR = \frac{1}{N}\sum_{i=1}^{N} AR_{i} \nonumber\\
      & IoU = \frac{Area\ of\ Overlap}{Area\ of\ Union} \label{eqn-8}
\end{align}

The rail defect datasets are evaluated on several advanced models and compared with the metrics shown in Table 4. It should be emphasized that the size of the model is limited, with the number of value Flops around 200G and the number of model parameters at <50M, to guarantee that the scale of the model has no discernible influence on the evaluation result. The detection model based on Swin transformer achieved a better performance on various categories, benefitting from the hierarchical feature representation on small subjects.\\

In addition, system robustness was observed. In certain categories such as Gap / Dent in the RIII dataset and Shelling in the MUET dataset, some other models perform better than Swin transformer. However, the control group cannot provide stability to the overall multi-class detection project. The overall precision and recall in Swin Transformer are both higher than in those of other models, where only MUET got a lower effectiveness because of the fuzzy image presentation under high-speed conditions. \\

\begin{table}[h]
    \begin{adjustwidth}{-2.0in}{0in}
    \centering
    \caption{{\bf System performance of different base models over rail surface defect datasets.}}
    \label{tab:baseline_model_performance}
    {%
    \begin{tabular}{lcccccccc}
        \toprule
        \textbf{Dataset} & \multicolumn{3}{c}{\textbf{MUET}} & \multicolumn{3}{c}{\textbf{RIII}} &  \multicolumn{2}{c}{\textbf{Model Parameter}} \\
                      & \textbf{mAP.50} & \textbf{mAP.75} & \textbf{AR@100} & \textbf{mAP.50} & \textbf{mAP.75} & \textbf{AR@100} & \textbf{Flops(G)} & \textbf{Params(M)} \\
        \midrule
        \textbf{YOLOX}\cite{r58}     & 0.535          & 0.418          & 0.554          & 0.612          & 0.161          & 0.470          & 13.3         & 8.9          \\
        \textbf{Mask R-CNN}\cite{r59} & 0.628          & 0.468          & 0.656          & 0.782          & 0.347          & 0.576          & 201          & 44.1         \\
         \textbf{DCNv2}\cite{r60}     & 0.612          & 0.477          & 0.647          & 0.770          & 0.321          & 0.558          & 182          & 44.9         \\
         \textbf{ResNet}\cite{r61}    & 0.639          & 0.541          & 0.656          & 0.780          & 0.380          & 0.566          & 201          & 44.1         \\
         \textbf{Swin\_T}\cite{r30}   &  \textbf{0.642}         &  \textbf{0.552}          &  \textbf{0.651}          &  \textbf{0.813}          &  \textbf{0.436}          & \textbf{0.573}          & \textbf{196}          & \textbf{47.5}         \\
        \bottomrule
    \end{tabular}%
    }
    \vspace{1em}
    \end{adjustwidth}
\end{table}

Compared to other object detection models, particularly the Yolo series models that prioritize speed, Swin transformer, which is based on the two-stage theory, has superior precision for the rail surface defect datasets. Ordinary Swin Transformer presents acceptable detection results for rail system defects like Gap and Crack in RIII. While for other categories like Dirt and Dent, considering the size of the instance and high degree of resemblance to the image background, the accuracy is slightly lower than average. \\

Rail track connection is symbolized by the Gap and Joint groups. They are not considered defects yet long-term monitoring is required since stress concentration caused by high-frequency collisions between wheels and rail at these positions may be catastrophic. Swin Transformer has a strong performance for Gap in the RIII dataset and Joint in the MUET dataset, and it can also identify Crack and Crush with high precision. This is because, despite the small instance size, the variation between smooth rails and instance edges is significant enough to be accurately detected. Other categories, such as Damage and Shelling, which are surface defects caused by friction, are more difficult to detect given their high resemblance with the background. Swin Transformer thus needs to be enhanced to achieve better performance. \\

\begin{table}[h]
    \begin{adjustwidth}{-2.0in}{0in}
    \centering
    \caption{{\bf System performance of different CBAM enhanced methods.}}
    \label{tab:CBAM_peformance}
    \resizebox{1.4\columnwidth}{!}
    {%
    \begin{tabular}{lcccccccc}
        \toprule
        \textbf{Model} & \multicolumn{3}{c}{\textbf{MUET}} & \multicolumn{3}{c}{\textbf{RIII}} & \multicolumn{2}{c}{\textbf{Log Analysis}} \\
                      & \textbf{mAP.50} & \textbf{mAP.75} & \textbf{AR@100} & \textbf{mAP.50} & \textbf{mAP.75} & \textbf{AR@100} & \textbf{Average iter time(s/iter)} & \textbf{Time std} \\
        \midrule
        Swin\_T       & 0.642          & 0.442          & 0.651          & 0.813          & 0.401          & 0.573          & 0.18±0.01      & 0.006 \\
        CBAM-SwinT-ML & 0.643          & 0.448          & 0.634          & 0.836          & 0.430          & 0.580          & 0.19±0.01      & 0.008 \\
        CBAM-SwinT-SL & 0.654          & 0.457          & 0.655          & 0.850          & 0.453          & 0.607          & 0.19±0.02      & 0.006 \\
        CBAM-SwinT-BL &  \textbf{0.691}          &  \textbf{0.543}          &  \textbf{0.690}          &  \textbf{0.881}          &  \textbf{0.490}         &  \textbf{0.581}          &  \textbf{0.22±0.02}      & \textbf{0.01}  \\
        \bottomrule
    \end{tabular}
    }
    \vspace{1em}
    \end{adjustwidth}
\end{table}

Table 5 illustrates how CBAM improves model overall performance. Average Precision and Recall improve about 5\% in each dataset, indicating CBAM module works on a better system performance. \\

Given the overlapping nature of the Swin Transformer windows, feature information which were not considered at the edge of the previous layer is emphasized in the current layer, to guarantee that the continuous layer framework can effectively capture the global context while constraining computational complexity to maintain linearity with the size of the input image. The features discussed above improve the effectiveness of the Swin Transformer over traditional CNN and Transformer models, but it is inefficient for detecting rail surface defects, especially with low light intensity and small object sizes. CBAM can dynamically fine-tune model features, reducing redundancy and improving feature diversity. The spatial attention and channel attention modules can substantially enhance feature maps and give concentrated supplementation for the entire system. \\

The system performance improvement depends on the position of CBAM allocated in the model. According to the experiment, model performance can be enhanced significantly if the CBAM module is in the deep level of the transformer framework close to the multi-head-attention, while the effect is lessened when the module is positioned outside the transformer model directly connecting to the input image. It is because the CBAM module’s limited capacity to handle high-resolution images. The hierarchical layer structure in Swin Transformer divides input features into several overlapping small sections, which enable CNN based module to work more efficiently.

\begin{figure}
    \begin{adjustwidth}{-2.25in}{0in}
    \centering
    \includegraphics[width=1\linewidth]{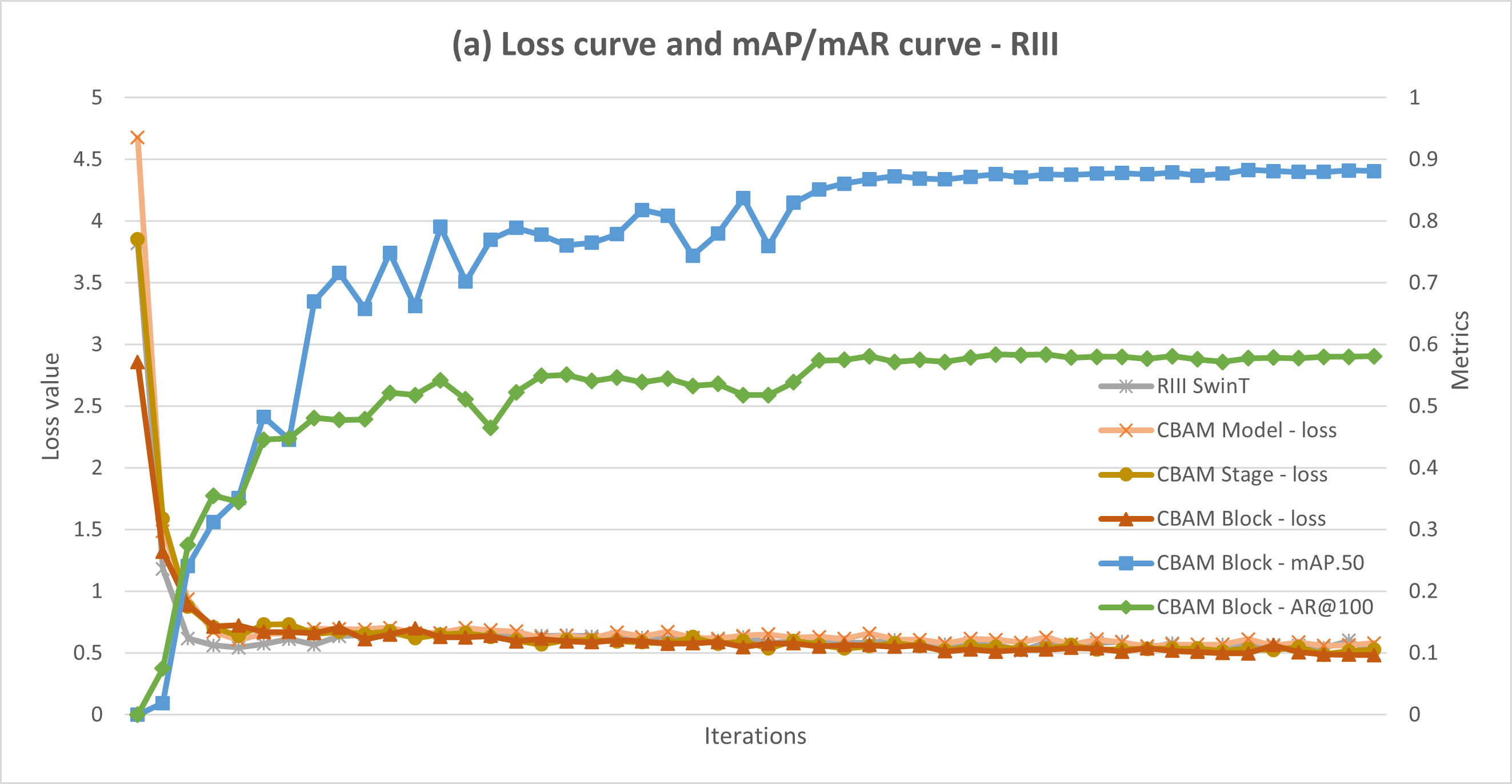}
    \includegraphics[width=1\linewidth]{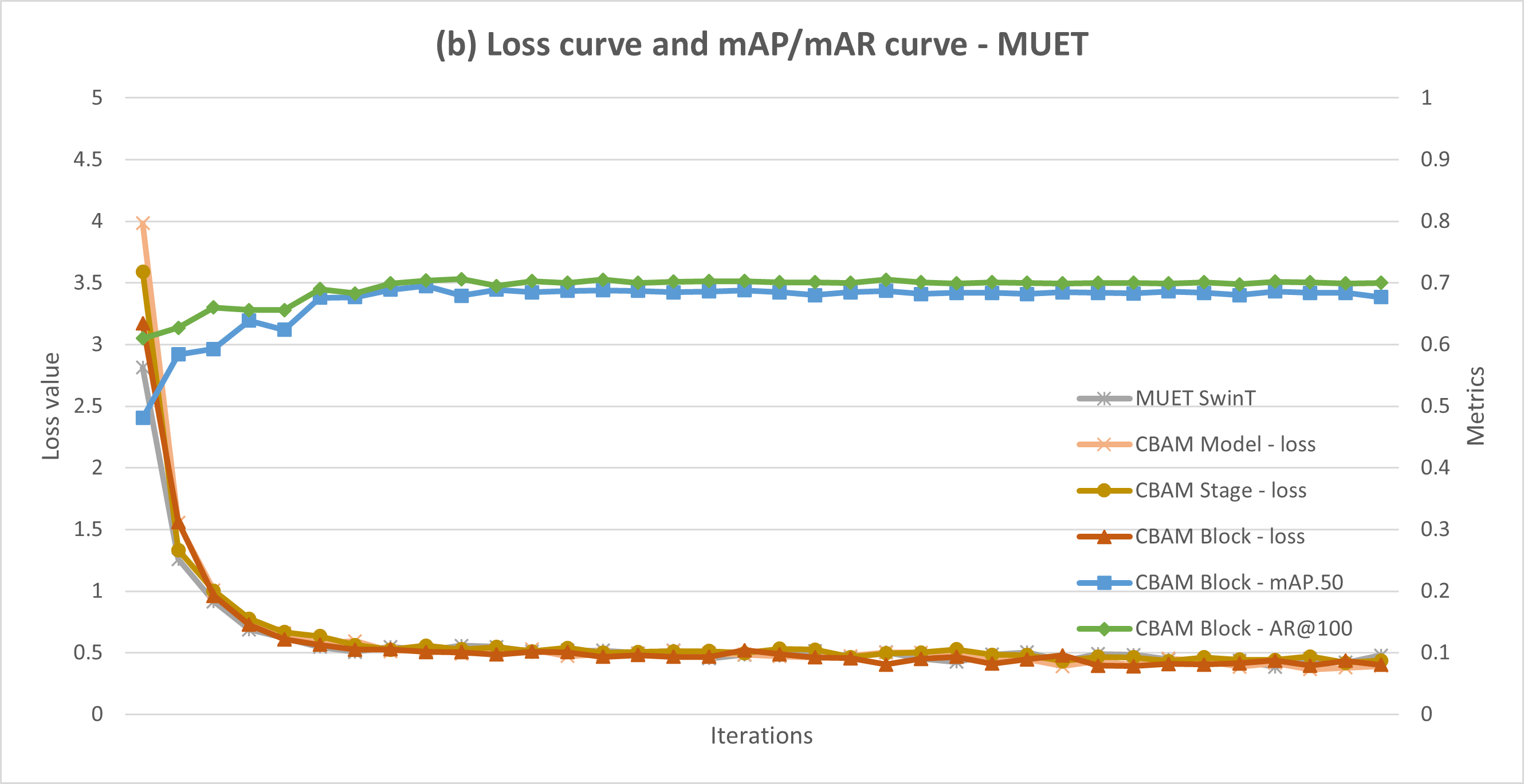}
    \caption{{\bf Loss curve and mAP/mAR curve of rail surface defect dataset.} 
    (a) RIII dataset. (b) MUET dataset.}
    \label{fig:Loss_curve}
    \end{adjustwidth}
\end{figure}

Figure 10 shows the loss value by iterations, indicating that the additional modules do not have negative impact of the training process. Log analysis in Table 5 shows the average training time per iteration. The average training speed(s/iter) and its standard deviation(std) may be susceptible with the integration of external modules, while the model parameters and learning rate remain constant. CBAM-SwinT-BL takes longer to be trained in each iteration compared to CBAM-SwinT-ML. The model depth is [2, 2, 6, 2] based on Swin Transformer configuration, which means that, at the model level, CBAM additional calculation perform once, while at the block level, CAM and SAM together operates 12 computations every iteration. Therefore, the convergence rate of the algorithm and the model’s stability slightly slowed down during model training with CBAM-SwinT-BL. However, considering the improvement of the system performance, the decrease in training speed is considered acceptable. \\

\subsection*{Detection Results Visualization}

The detection results for the CBAM-enhanced Swin Transformer in the rail surface defect dataset are shown in Figure 11. The figure contains mAP-50 of each category, arranged by the size of instances compared to the whole image, which has been described as column \(Avg\_size by Ratio\) in Table 1. The horizontal axis is arranged in descending order by size ratio. \\

\begin{figure}
    \begin{adjustwidth}{-2.25in}{0in}
    \centering
    \includegraphics[width=1\linewidth]{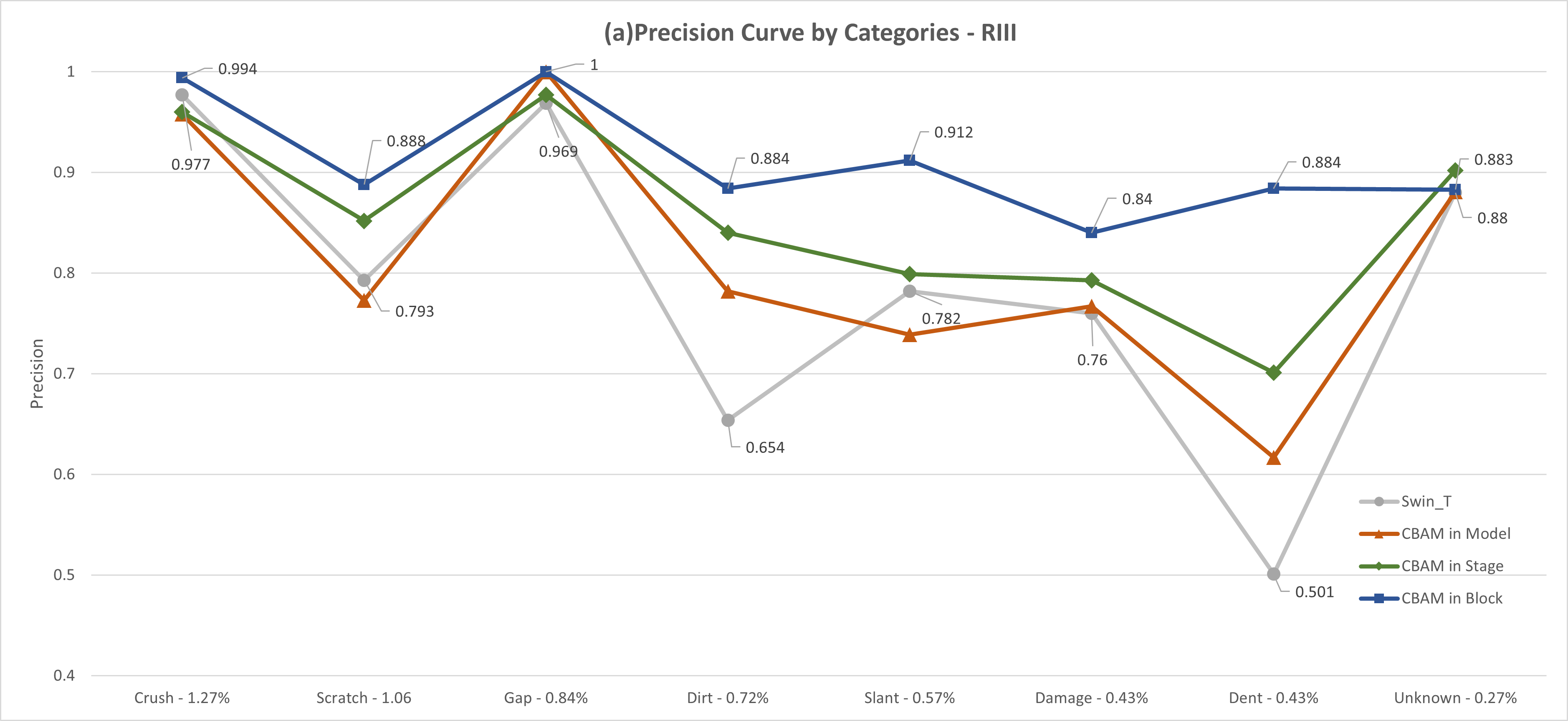}
    \includegraphics[width=1\linewidth]{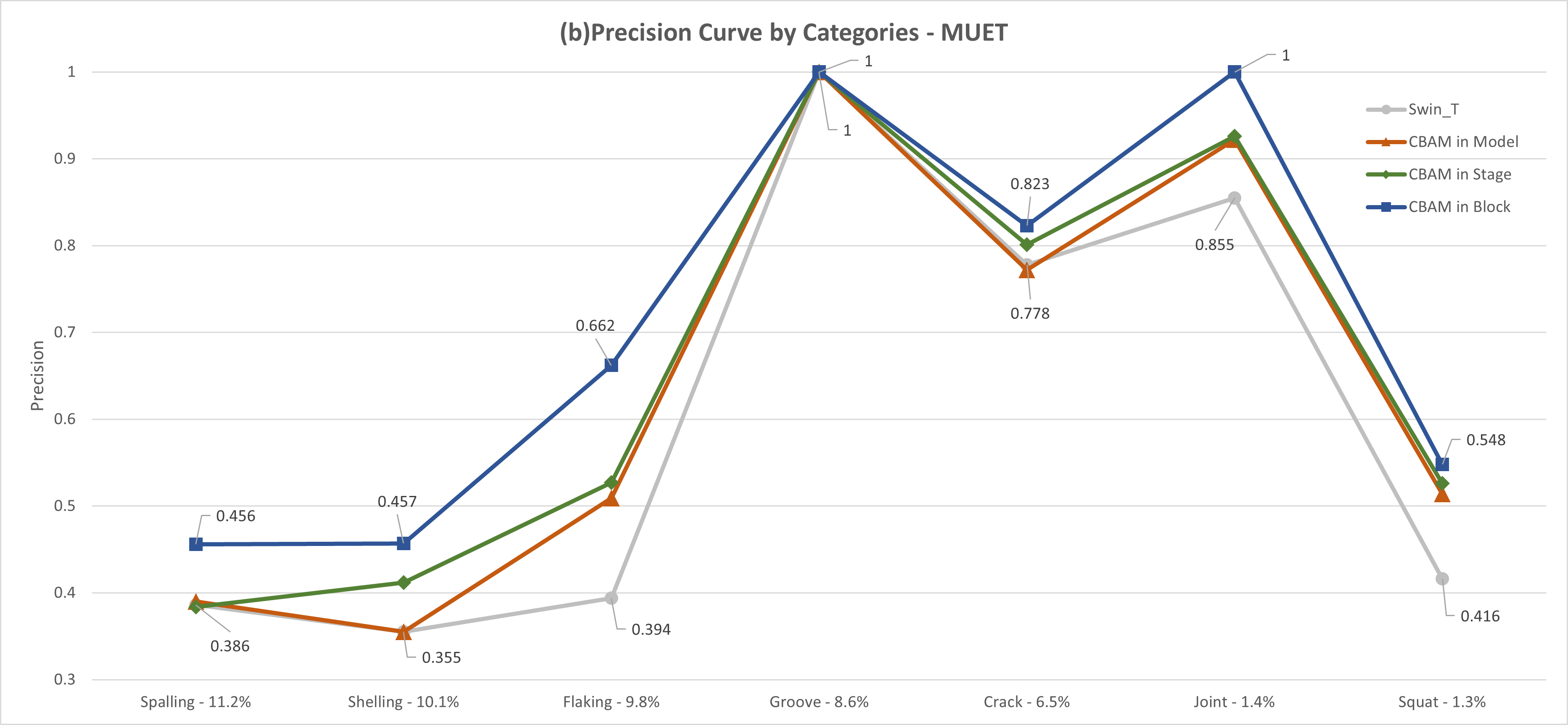}
    \caption{{\bf Precision curve by category in rail surface dataset, followed by instance size ratio compared to the overall image.} (a) RIII dataset. (b) MUET dataset.}
    \label{fig:RIII_precision_curve}
    \label{fig:MUET_precision_curve}
    \end{adjustwidth}
\end{figure}

The comparison reveals a considerable difference between the detection accuracy of the original Swin Transformer model and the CBAM-improved model in Figure 11. In several categories with smaller sizes, such as Dirt, Slant, Damage, and Dent, the CBAM-SwinT-BL significantly improved the overall precision by 23\%, 13\%, 12\%, and 38.4\%, respectively.\\

Focusing on the low precision of the original Swin Transformer on Dirt and Dent in the RIII dataset, it is noticed that some images are detected as Damage during model training, which is because they all resemble scratches on the rail surface visually and are difficult for Swin Transformer to differentiate. With the CBAM module, Dent instances can be more accurately detected, due to their smaller size, sharper object boundary, and more obvious dark fringe, compared to Dirt instances. For Slant instances, due to the narrow size of slants and their low contrast with the background rail surface environment, accurate detection is more difficult compared with other classes. Yet, with image enhancement, results with higher precision can be obtained, which is now steady at around 85\%. \\

In contrast with traditional railroad images taken within tunnels, the images from the MUET dataset are outdoor and the tracks' surface is prone to significant deterioration. Among the defects, Shelling, spalling, and flaking conflict with one another for comparable surface damage, which is an important attribute contributing to the MUET dataset's inadequate performance. Hence, it is essential for research to acknowledge the challenge of implementing detection models in various situations. Two datasets also show various data quality with different railway conditions. For instance, the Joint class in the MUET dataset refers to the apparent distance between the rails at both ends, while the Gap class in the RIII dataset refers to the seamless connection. The joint class in MUET achieves a precision increase by 15\% with CBAM-enhanced applications whereas the Gap class in RIII can already be accurately detected with an ordinary detection model. Meanwhile, improving the performance of blurry image detection remains an essential topic that needs to be resolved. \\

\begin{figure}
\begin{adjustwidth}{-2.25in}{0in}
    \centering
    \includegraphics[width=1\linewidth]{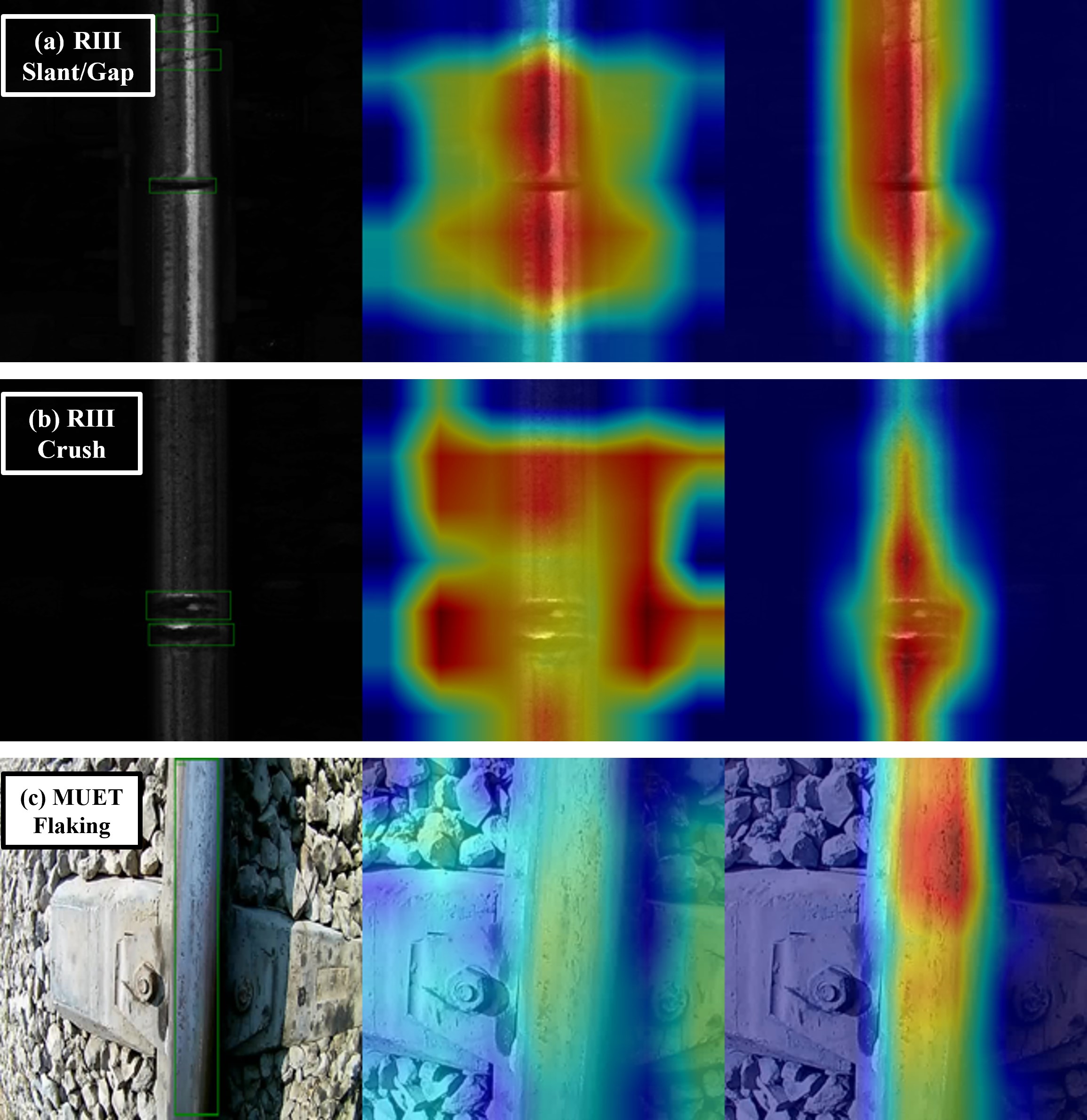}
    \caption{{\bf Model attention maps for RIII and MUET dataset.} Left column is the origin image with bbox, middle column is SwinT model and right column is the CBAM-SwinT-BL method. 
    (a) slant and gap in RIII. (b) crush. (c) flaking in MUET.}
    \label{fig:CAM}
    \end{adjustwidth}
\end{figure}

Attention maps in Figure 12 indicating that CBAM module has a significant concentration of model attention on target area shown in the left image. While the transformer-based methods producing good visual attention maps in all three cases, in the absence of the CBAM enhancement approach, it is more susceptible to background disturbance. In figure 12 (a) the right image, the attention map covers the entire rail, ensuring that the slants at the top of the image are properly recognized. In figure 12(b)(c), the attention focusing on the potential defect location, indicating the model can identify defects more accurately.\\

\begin{figure}
    \begin{adjustwidth}{-2.25in}{0in}
    \centering
    \includegraphics[width=1\linewidth]{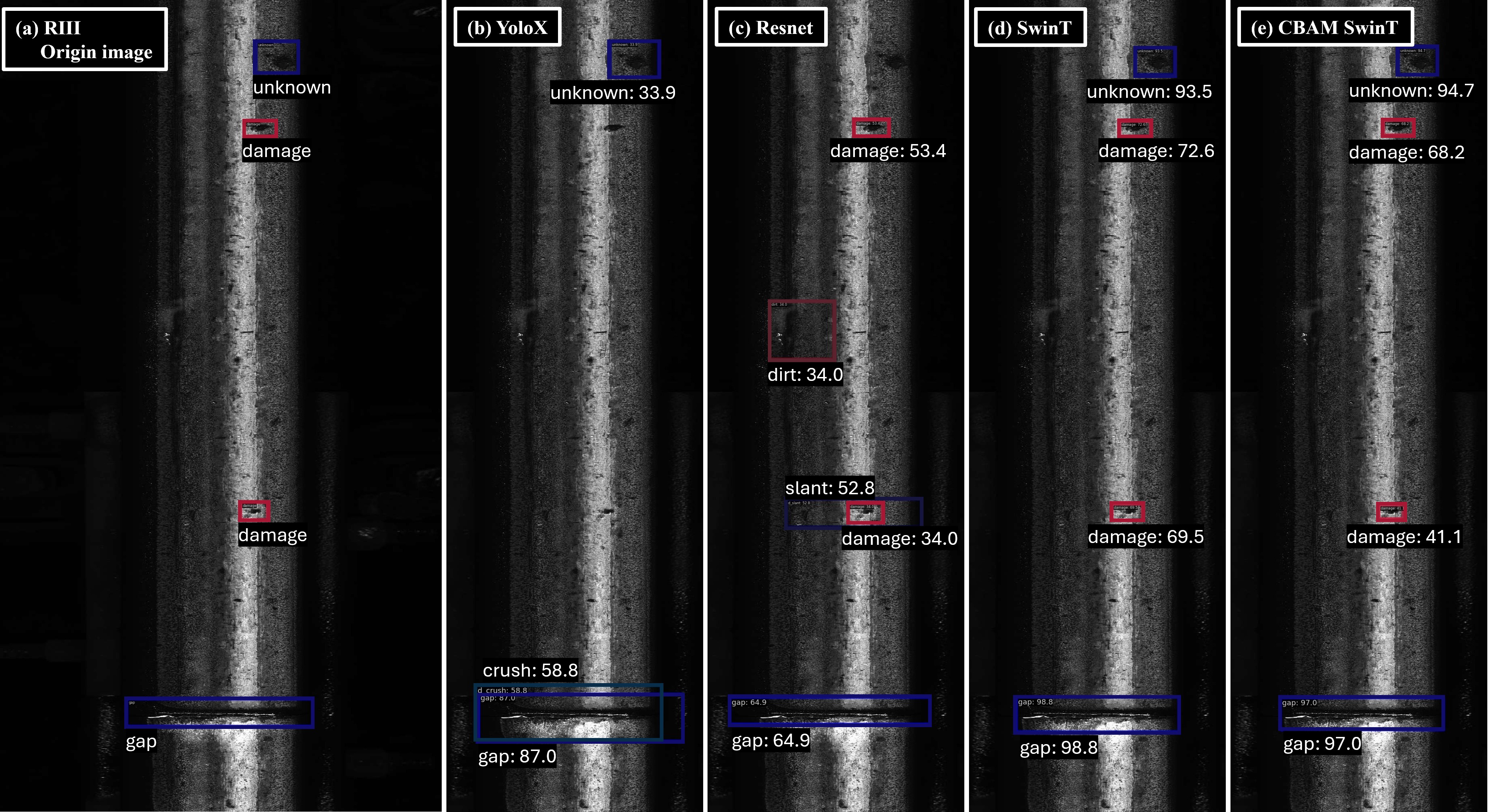}
    \caption{{\bf Detection result of RIII - unknown/damage/gap.} (a) original image. (b) YoloX. (c) Resnet. (d) simple SwinT. (e) CBAM-SwinT-BL.}
    \label{fig:Detection result of RIII - 1}
    \end{adjustwidth}
\end{figure}

\begin{figure}
    \begin{adjustwidth}{-2.25in}{0in}
    \centering
    \includegraphics[width=1\linewidth]{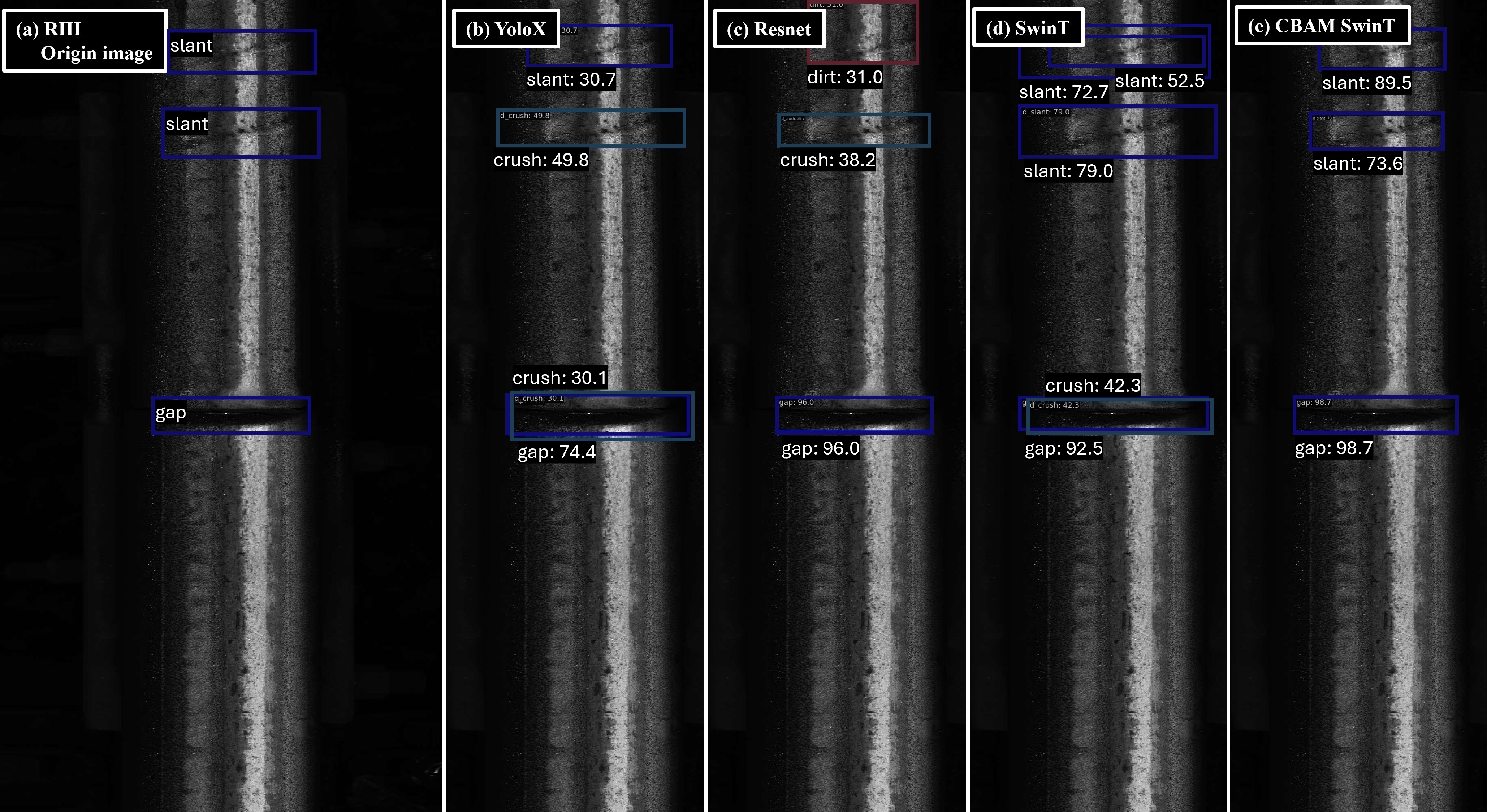}
    \caption{{\bf Detection result of RIII - slant /gap.} (a) original image. (b) YoloX. (c) Resnet. (d) simple SwinT. (e) CBAM-SwinT-BL.}
    \label{fig:Detection result of RIII - 2}
    \end{adjustwidth}
\end{figure}

\begin{figure}
    \begin{adjustwidth}{-2.25in}{0in}
    \centering
    \includegraphics[width=1\linewidth]{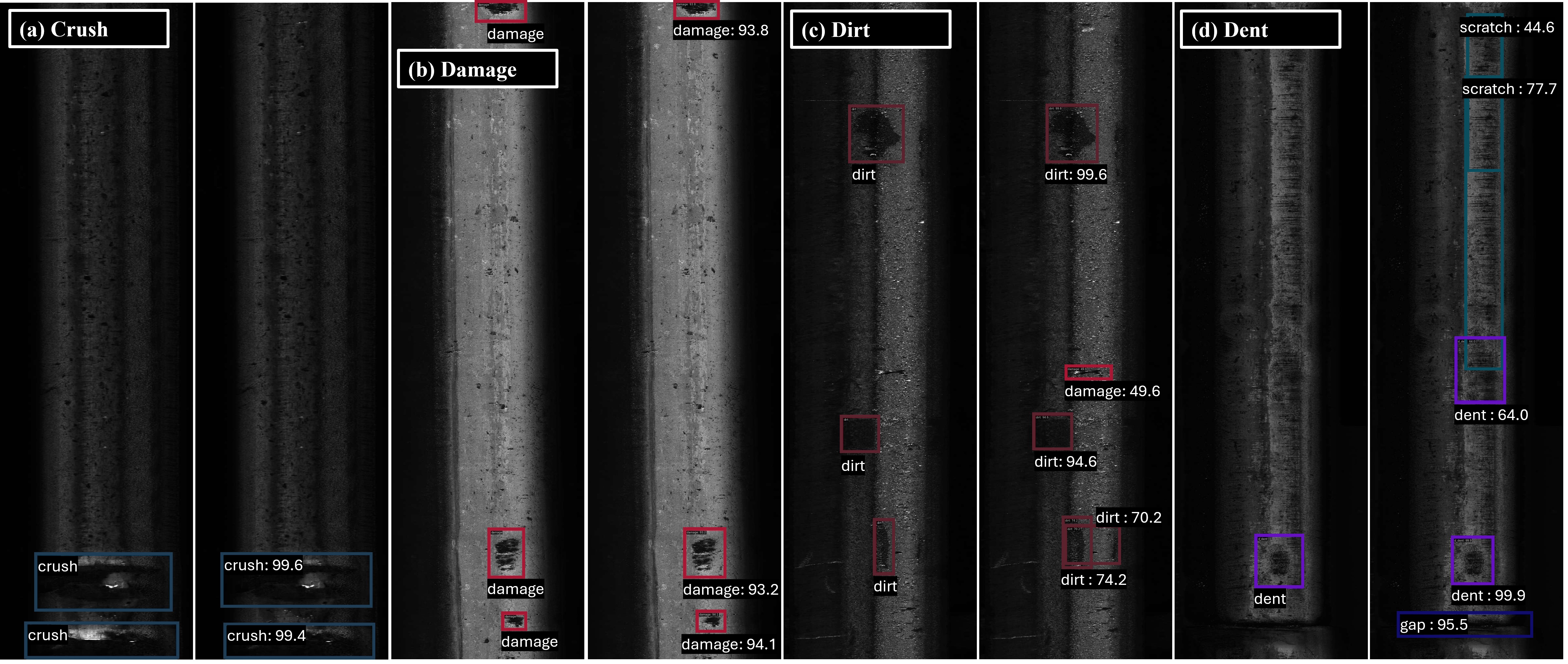}
    \caption{{\bf Detection result of RIII in CBAM-SwinT-BL.} (a) crush. (b) damage. (c) dirt. (d) dent.}
    \label{fig:Detection result of RIII - CBAM}
    \end{adjustwidth}
\end{figure}

Figure 13 and Figure 14 show the detection results of RIII dataset. It was shown that YoloX, a one-stage CNN-based, missed detection in the category Damage and misidentified Slant. This demonstrates that after many version iterations, the lightweight Yolo model can still be ineffective when it comes to detecting defects in certain scenarios. Attaining convincing outcomes is a difficult endeavour. The Mask-RCNN based resnet-strikes-back model and swin transformer model have acquired far better reliability compares to Yolox. Those models can effectively recognize an obvious part of defects and perform relatively good in identification and localization. However, it is difficult to identify defects with small effective areas or fuzzy boundaries such as Slant in Figure 14, and the detection result are not satisfied for the overall system. \\

Figure 16 displays comparable results. While CBAM-SwinT-BL still produces consistent detection output under low image quality conditions, YoloX and Resnet both exhibit false positives and false negatives for the identification and localization of category Squat and Joint. The MUET dataset was annotated manually which may have considerably reduced the quality as a training set. However, the detection result of CBAM-SwinT-BL for Squat is more accurate than that in manual annotation, indicating that the detection model is effective for the real operation environment. \\

\begin{figure}
    \begin{adjustwidth}{-2.25in}{0in}
    \centering
    \includegraphics[width=1\linewidth]{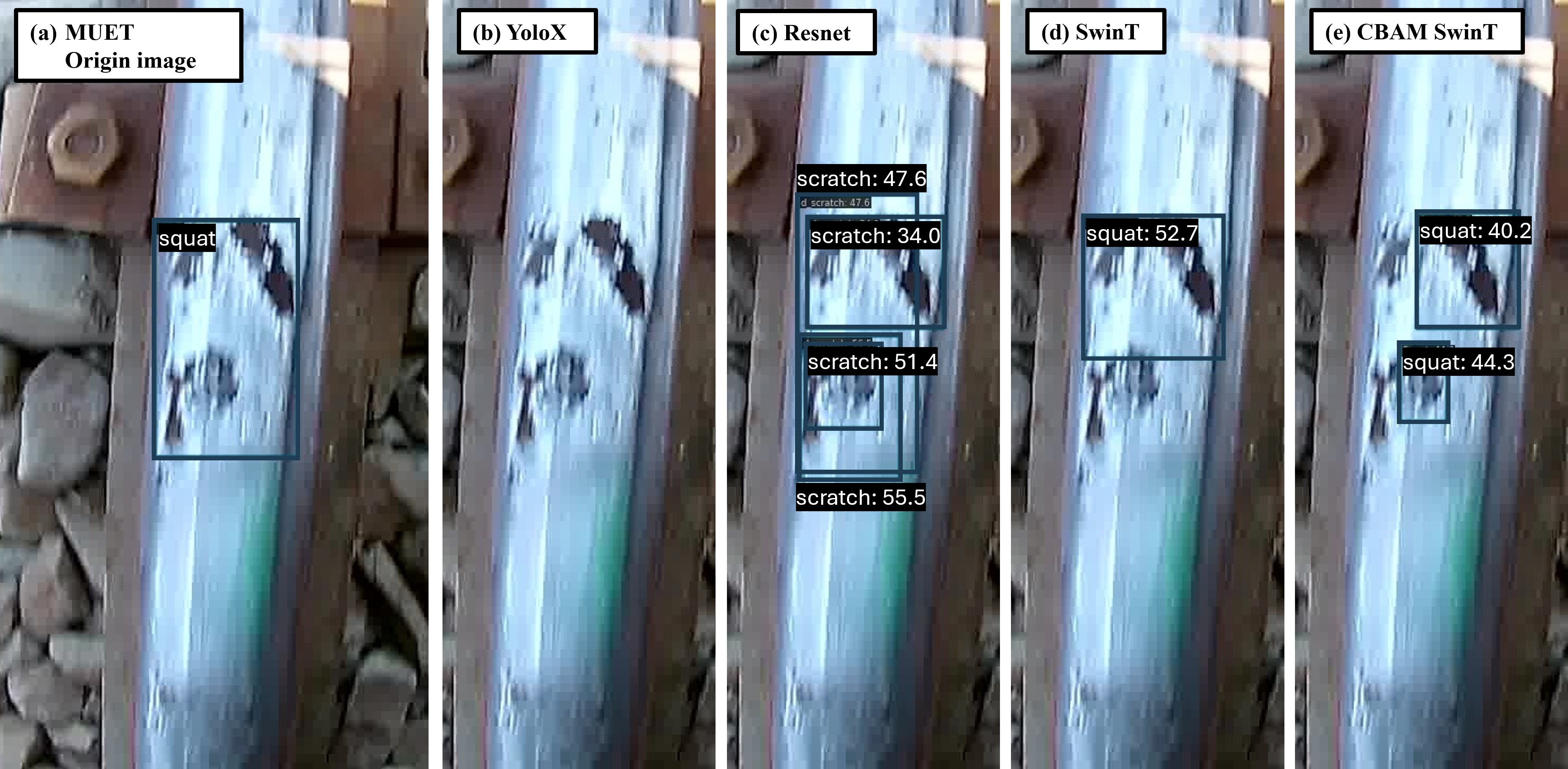}
    \caption{{\bf Detection result of MUET - squat.} (a) original image. (b) YoloX. (c) Resnet. (d) simple SwinT. (e) CBAM-SwinT-BL.}
    \label{fig:Detection result of MUET - 1}
    \end{adjustwidth}
\end{figure}

\begin{figure}
    \begin{adjustwidth}{-2.25in}{0in}
    \centering
    \includegraphics[width=1\linewidth]{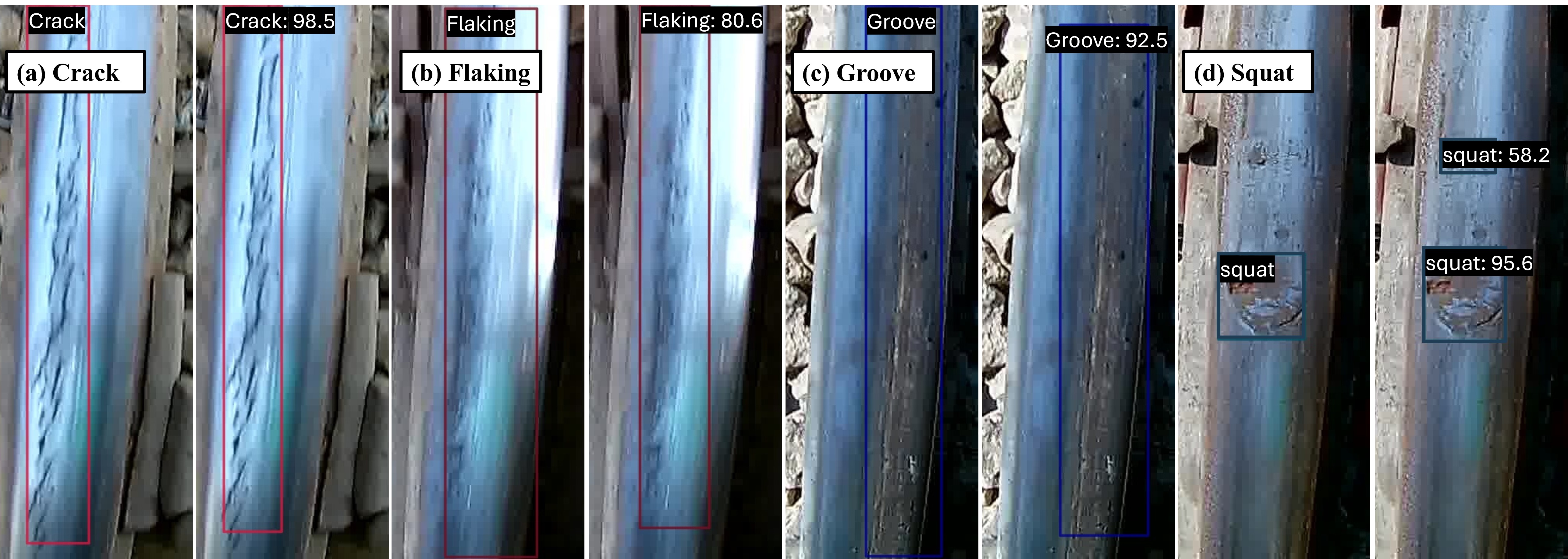}
    \caption{{\bf Detection result of MUET in CBAM-SwinT-BL.} (a) crack. (b) flaking. (c) groove. (d) squat.}
    \label{fig:MUET_CBAM_vis}
    \end{adjustwidth}
\end{figure}

Some defect edges in the MUET dataset are unclear because of the image capture equipment's limitations, which will negatively affect the model's inference bounding box area, and several defects in the high-resolution, low-contrast RIII dataset images are still difficult to identify even after image enhancement. CBAM-enhanced Swin Transformer, on the other hand, has proved its consistently reliable accuracy on both MUET and RIII datasets even for small defects with fuzzy instance edges in the ablation study. Those aforementioned challenges are essentially solved by the Block-level CBAM-enhanced Swin transformer shown in Table 5 and Figure 13, 14 and 16.\\

The model precisely detects each piece of defects including Crush, Damage, Dirt and Dent in Figure 15, and it sensitively recognizes the surface wear like Crack, Flaking, Groove and Squat in Figure 17. The false positive alert, for example “damage” in Figure 15(c) and “dent, scratch” in Figure 15(d) along with “squat” in Figure 17(d) may reduce the accuracy of the system slightly. However, for a railway maintenance system which may consider safety as its highest priority, is acceptable to sacrifice a small degree of accuracy as a trade-off for a higher sensitivity. The outputs of the model for the identification of fuzzy borders have been shown in Figure 17(a, b, c). It is obvious that the detection model with CBAM-SwinT-BL is capable of recognizing defects with blurred edges in broad regions, indicating that the model can also be helpful in detecting surface defects under equipment limitations.

\section*{Conclusion}

With the development of computer vision algorithm, object detection model has been developed to automatically detect rail surface defects in railway maintenance system. This paper investigates a method focusing on small object detection on rail defects where such categories, such as Dirt and Squat, have a relative size <2\%. The proposed method is based on CBAM-enhanced Swin Transformer algorithm. Model evaluation were evaluated on public datasets RIII and MUET, which were annotated on both normal rail system components and rail surface defects. This serves as a research foundation for computer vision-based autonomous railway management systems. \\

The experiment began with a comparison of several baseline object detection models based on CNN and Transformer. The experiment result shows that Swin Transformer reaches a higher accuracy and stability due to the Two-Stage framework’s benefits in feature extraction and object localization. To reduce the effect of low-light environments in railway tunnels and data imbalances between categories, data preprocess was employed including data augmentation and image enhancement. The proposed method utilized an innovative combination of the Swin Transformer and the Convolutional-Based Attention Module to optimize small-area targets and demonstrate that block-level integration can be the most optimal framework. \\

The experiment and ablation study was processed on RIII and MUET datasets, where CBAM-SwinT-BL outperformed the standard model by 6.8\% and 4.9\% on average precision, reaching 88.1\% and 69.1\%, respectively, indicating the efficient of proposed method. The mAP-50 improvement on RIII-Dirt and MUET-Squat also demonstrate that CBAM-SwinT-BL can achieve performance significantly higher than ordinary model. The various CBAM and Swin Transformer combination approaches reveal that the attention module has a better performance when it directly operates on the input vector that has been partitioned, making the entire system more accurate in identifying small-size defects. This research makes a notably contribution to computer vision-based rail defect detection and motivates future work to improve the safety of railway system. \\

\section*{Acknowledgments}
The work presented in this article is supported by the Centre for Advances in Reliability and Safety (CAiRS) admitted under AiR@InnoHK Research Cluster.


%
%
%

\end{document}